\documentclass{article}

\usepackage{microtype}
\usepackage{graphicx}
\usepackage{subfigure}
\usepackage{booktabs} 

\usepackage{hyperref}

\usepackage[accepted]{icml2019}

\icmltitlerunning{Understanding MCMC Dynamics as Flows on the Wasserstein Space}

\usepackage[lower, theorems, autonum, eqref]{changdefs}
\usepackage{mathtools} 
\usepackage{wrapfig}
\newcommand{\clCC}{\clC_c^{\infty}}
\newcommand{\tclP}{{\widetilde \clP}}
\DeclareMathOperator{\gradf}{grad_{fib}}
\newtheorem{assumption}[theorem]{Assumption}

\usepackage{titlesec}
\titlespacing*{\section}{0pt}{0pt}{-4pt}
\titlespacing*{\subsection}{0pt}{0pt}{-4pt}
\titlespacing*{\subsubsection}{0pt}{0pt}{-4pt}

\usepackage{color}

\begin{document}
\abovedisplayskip=3pt
\belowdisplayskip=4pt
\abovedisplayshortskip=3pt
\belowdisplayshortskip=4pt

\twocolumn[
  \icmltitle{Understanding MCMC Dynamics as Flows on the Wasserstein Space}



\icmlsetsymbol{equal}{*}

\begin{icmlauthorlist}
  \icmlauthor{Chang Liu}{thu}
  \icmlauthor{Jingwei Zhuo}{thu}
  \icmlauthor{Jun Zhu}{thu}
\end{icmlauthorlist}

\icmlaffiliation{thu}{Dept. of Comp. Sci. \& Tech., Institute for AI, BNRist Center, Tsinghua-Fuzhou Inst. for Data Tech., THBI Lab, Tsinghua University, Beijing, 100084, China}

\icmlcorrespondingauthor{Jun Zhu}{dcszj@tsinghua.edu.cn}

\icmlkeywords{MCMC, particle-based variational inference, Wasserstein space, gradient flow, Hamiltonian flow, Bayesian inference}

\vskip 0.3in
]



\printAffiliationsAndNotice{}  

\begin{abstract}
  It is known that the Langevin dynamics used in MCMC is the gradient flow of the KL divergence on the Wasserstein space, which helps convergence analysis and inspires recent particle-based variational inference methods (ParVIs).
  But no more MCMC dynamics is understood in this way.
  In this work, by developing novel concepts, we propose a theoretical framework that recognizes a general MCMC dynamics as the fiber-gradient Hamiltonian flow on the Wasserstein space of a fiber-Riemannian Poisson manifold.
  The ``conservation + convergence'' structure of the flow gives a clear picture on the behavior of general MCMC dynamics.
  The framework also enables ParVI simulation of MCMC dynamics, which enriches the ParVI family with more efficient dynamics, and also adapts ParVI advantages to MCMCs.
  We develop two ParVI methods for a particular MCMC dynamics and demonstrate the benefits in experiments.
\end{abstract}

\section{Introduction}
Dynamics-based Markov chain Monte Carlo methods (MCMCs) in Bayesian inference have drawn great attention because of their wide applicability, efficiency, and scalability for large-scale datasets \cite{neal2011mcmc, welling2011bayesian, chen2014stochastic, chen2016stochastic, li2019communication}.
They draw samples by simulating a continuous-time \emph{dynamics}, or more precisely, a diffusion process, that keeps the target distribution invariant.
However, they often exhibit slow empirical convergence and relatively small effective sample size, due to the positive auto-correlation of the samples.
Another type of inference methods, called particle-based variational inference methods (ParVIs), aim to deterministically update samples, or particles as they call them, so that the particle distribution minimizes the KL divergence to the target distribution.
They fully exploit the approximation ability of a set of particles by imposing an interaction among them, so they are more particle-efficient.
Optimization-based principle also makes them convergence faster.
Stein variational gradient descent (SVGD) \cite{liu2016stein} is the most famous representative, and the field is under an active development both in theory \cite{liu2017steinflow, chen2018stein, chen2018unified, liu2019understanding_a} and application \cite{liu2017steinpolicy, pu2017vae, zhuo2018message, yoon2018bayesian}.

The study on the relation between the two families starts from their interpretations on the Wasserstein space $\clP(\clM)$ supported on some smooth manifold $\clM$ \cite{villani2008optimal, ambrosio2008gradient}.
It is defined as the space of distributions
\begin{align}
  \clP(\clM) := \{ q \mid\; & \text{$q$ is a probability measure on $\clM$ and} \notag\\
  & \exists x_0 \in \clM \st \bbE_{q(x)}[d(x,x_0)^2] < \infty \}
  \label{eqn:p2}
\end{align}
with the well-known Wasserstein distance. 
It is very general yet still has necessary structures.
With its canonical metric, the gradient flow (steepest descending curves) of the KL divergence is defined.
It is known that the Langevin dynamics (LD) \cite{langevin1908theorie, roberts1996exponential}, a particular type of dynamics in MCMC, simulates the gradient flow on $\clP(\clM)$ \cite{jordan1998variational}.
Recent analysis reveals that existing ParVIs also simulate the gradient flow \cite{chen2018unified, liu2019understanding_a}, so they simulate the same dynamics as LD.
However, besides LD, there are more types of dynamics in the MCMC field that converge faster and produce more effective samples \cite{neal2011mcmc, chen2014stochastic, ding2014bayesian}, but no ParVI yet simulates them.
These more general MCMC dynamics have not been recognized as a process on the Wasserstein space $\clP(\clM)$, and this poses an obstacle towards ParVI simulations.
On the other hand, the convergence behavior of LD becomes clear when viewing LD as the gradient flow of the KL divergence on $\clP(\clM)$ (\eg, \citet{cheng2017convergence}), which leads distributions to the target in a steepest way.
However, such knowledge on other MCMC dynamics remains obscure, except a few.
In fact, a general MCMC dynamics is only guaranteed to keep the target distribution invariant \cite{ma2015complete}, but unnecessarily drives a distribution towards the target steepest.
So it is hard for the gradient flow formulation to cover general MCMC dynamics.

In this work, we propose a theoretical framework that gives a unified view of general MCMC dynamics on the Wasserstein space $\clP(\clM)$.
We establish the framework by two generalizations over the concept of gradient flow towards a wider coverage:
\textbf{(a)} we introduce a novel concept called \emph{fiber-Riemannian manifold} $\clM$, where only the Riemannian structure on each fiber (roughly a decomposed submanifold, or a slice of $\clM$) is required, and we develop the novel notion of \emph{fiber-gradient flow} on its Wasserstein space $\clP(\clM)$;
\textbf{(b)} we also endow a Poisson structure to the manifold $\clM$ and exploit the corresponding Hamiltonian flow on $\clP(\clM)$.
Combining both explorations, we define a fiber-Riemannian Poisson (fRP) manifold $\clM$ and a fiber-gradient Hamiltonian (fGH) flow on its Wasserstein space $\clP(\clM)$.
We then show 
that any regular MCMC dynamics is the fGH flow on the Wasserstein space $\clP(\clM)$ of an fRP manifold $\clM$, and there is a correspondence between the dynamics and the structure of the fRP manifold $\clM$.

This unified framework gives a clear picture 
on the behavior of MCMC dynamics.
The Hamiltonian flow conserves the KL divergence to the target distribution, while the fiber-gradient flow minimizes it on each fiber, driving each conditional distribution to meet the corresponding conditional target.
The target invariant requirement is recovered in which case the fiber-gradient is zero, and moreover, we recognize that the fiber-gradient flow acts as a stabilizing force on each fiber.
It enforces convergence fiber-wise, making the dynamics in each fiber robust to simulation with the noisy stochastic gradient, which is crucial for large-scale inference tasks.
This generalizes the discussion of \citet{chen2014stochastic} and \citet{betancourt2015fundamental} on Hamiltonian Monte Carlo (HMC) \cite{duane1987hybrid, neal2011mcmc, betancourt2017conceptual} to general MCMCs.
In our framework, different MCMCs correspond to different fiber structures and flow components.
They can be categorized into three types, each of which has its particular behavior.
We make a unified study on various existing MCMCs under the three types.

Our framework also bridges the fields of MCMCs and ParVIs, so that on one hand, the gate to the reservoir of MCMC dynamics is opened to the ParVI family and abundant efficient dynamics are enabled beyond LD, and on the other hand, MCMC dynamics can be now simulated in the ParVI fashion, inheriting advantages like particle-efficiency.
To demonstrate this, we develop two ParVI simulation methods for the Stochastic Gradient Hamiltonian Monte Carlo (SGHMC) dynamics \cite{chen2014stochastic}.
We show the merits of using SGHMC dynamics over LD in the ParVI field, and ParVI advantages over conventional stochastic simulation in MCMC.

\textbf{Related work}~~
\citet{ma2015complete} give a complete recipe on general MCMC dynamics.
The recipe guarantees the target invariant principle, but leaves the behavior of these dynamics unexplained.
Recent analysis towards a broader kind of dynamics via the Fokker-Planck equation \cite{kondratyev2017nonlinear, bruna2017asymptotic} is still within the gradient flow formulation, thus not general enough.

On connecting MCMC and ParVI, \citet{chen2018unified} explore the correspondence between LD and Wasserstein gradient flow, and develop new implementations for dynamics simulation.
However, their consideration is still confined on LD, leaving more general MCMC dynamics untouched.
\citet{gallego2018stochastic} formulate the dynamics of SVGD as a particular kind of MCMC dynamics, but no existing MCMC dynamics is recognized as a ParVI.
More recently, \citet{taghvaei2018accelerated} derive an accelerated ParVI that is similar to one of our ParVI simulations of SGHMC.
The derivation does not utilize the dynamics and the method connects to SGHMC only algorithmically.
Our theory solidates our ParVI simulations of SGHMC, and enables extensions to more dynamics.

\section{Preliminaries}
\label{sec:pre}

We first introduce the recipe for general MCMC dynamics~\cite{ma2015complete}, and prior knowledge on flows on a smooth manifold $\clM$ and its Wasserstein space $\clP(\clM)$. 

A smooth manifold $\clM$ is a topological space that locally behaves like an Euclidean space.
Since the recipe describes a general MCMC dynamics in an Euclidean space $\bbR^M$, it suffices to only consider $\clM$ that is globally diffeomorphic to $\bbR^M$, which is its global coordinate system.
For brevity we use the same notation for a point on $\clM$ and its coordinates due to their equivalence.
A tangent vector $v$ at $x \in \clM$ can be viewed as the differentiation along the curve that is tangent to $v$ at $x$, so $v$ can be expressed as the combination $v = \sum_{i=1}^M v^i \partial_i$ of the differentiation operators $\{ \partial_i := \frac{\partial}{\partial x^i} \}_{i=1}^M$, which serve as a set of basis of the tangent space $T_x \clM$ at $x$.
The cotangent space $T^*_x \clM$ at $x$ is the dual space of $T_x \clM$, and the cotangent bundle is the union $T^* \! \clM := \bigcup_{x \in \clM} T^*_x \clM$.
We adopt Einstein convention to omit the summation symbol for a pair of repeated indices in super- and sub-scripts (\eg, $v = v^i \partial_i := \sum_{i=1}^M v^i \partial_i$). 
We assume the target distribution to be absolutely continuous 
so that we have its density function $p$.

\subsection{The Complete Recipe of MCMC Dynamics}
\label{sec:pre-mcmc}

The fundamental requirement on MCMCs is that the target distribution $p$ is kept stationary under the MCMC dynamics.
\citet{ma2015complete} give a general recipe for such a dynamics expressed as a diffusion process in an Euclidean space $\bbR^M$:
\vspace{-10pt}
\begin{equation}\begin{gathered}
  \ud x = V(x) \dd t + \sqrt{2 D(x)} \dd B_t(x),\\
  V^i(x) = \frac{1}{p(x)} \partial_j \Big( p(x) \big( D^{ij}(x) + Q^{ij}(x) \big) \Big),
  \label{eqn:recipe}
\end{gathered}\end{equation}
for any positive semi-definite matrix $D_{M \times M}$ (diffusion matrix) and any skew-symmetric matrix $Q_{M \times M}$ (curl matrix),
where $B_t(x)$ denotes the standard Brownian motion in $\bbR^M$. 
The term $V(x) \dd t$ represents a deterministic drift and $\sqrt{2 D(x)} \dd B_t(x)$ a stochastic diffusion.
It is also shown that if $D$ is positive definite, $p$ is the unique stationary distribution.
Moreover, the recipe is complete, \ie, any diffusion process with $p$ stationary can be cast into this form.

The recipe gives a universal view and a unified way to analyze MCMCs.
In large scale Bayesian inference tasks, the stochastic gradient (SG), a noisy estimate of $(\partial_j \log p)$ on a randomly selected data mini-batch, is crucially desired for data scalability.
The dynamics is compatible with SG, since the variance of the drift is of higher order of the diffusion part \cite{ma2015complete, chen2015convergence}.
In many MCMC instances, $x = (\theta, r)$ is taken as an augmentation of the target variable $\theta$ by an auxiliary variable $r$. 
This could encourage the dynamics to explore a broader area to reduce sample autocorrelation and improve efficiency (\eg, \citet{neal2011mcmc, ding2014bayesian, betancourt2017geometric}).

\subsection{Flows on a Manifold}
\label{sec:pre-flow}

The mathematical concept of the \emph{flow} associated to a vector field $X$ on $\clM$ is a set of curves on $\clM$, $\{ (\varphi_t(x))_t \mid x \in \clM \}$, such that the curve $(\varphi_t(x))_t$ through point $x \in \clM$ satisfies $\varphi_0(x) = x$ and that its tangent vector at $x$, $\frac{\ud}{\ud t} \varphi_t(x) \big|_{t=0}$, coincides with the vector $X(x)$.
For any vector field, its flow exists at least locally (\citet{do1992riemannian}, Sec.~0.5).
We introduce two particular kinds of flows for our concern.

\subsubsection{Gradient Flows}
\label{sec:pre-flow-grad}

We consider the gradient flow on $\clM$ induced by a \emph{Riemannian structure} $g$ (\eg, \citet{do1992riemannian}), which gives an inner product $g_x(\cdot,\cdot)$ in each tangent space $T_x \clM$. 
Expressed in coordinates, $g_x(u,v) = g_{ij}(x) u^i v^j, \forall u = u^i \partial_i, v = v^i \partial_i \in T_x \clM$, and the matrix $(g_{ij}(x))$ is required to be symmetric (strictly) positive definite.
The \emph{gradient} of a smooth function $f$ on $\clM$ can then be defined as the steepest ascending direction and has the coordinate expression: 
\begin{align}
  \grad f(x) = g^{ij}(x) \partial_j f(x) \partial_i \quad \in T_x \clM,
  \label{eqn:grad}
\end{align}
where $g^{ij}(x)$ is the entry of the inverse matrix of $(g_{ij}(x))$.
It is a vector field and determines a gradient flow.

On $\clP(\clM)$, a Riemannian structure is available with a Riemannian support $(\clM, g)$ \cite{otto2001geometry, villani2008optimal, ambrosio2008gradient}.
The tangent space at $q \in \clP(\clM)$ is recognized as (\citet{villani2008optimal}, Thm.~13.8; \citet{ambrosio2008gradient}, Thm.~8.3.1): 
\vspace{-3pt}
\begin{align}
  T_{q} \clP(\clM) = \overline{ \{ \grad f \mid f \in \clCC(\clM) \} }^{\clL^2_{q}(\clM)},
  \label{eqn:p2tg}
\end{align}
where $\clCC(\clM)$ is the set of compactly supported smooth functions on $\clM$, $\clL^2_{q}(\clM)$ is the Hilbert space $\{ \text{$X$: vector field on $\clM$} \mid \bbE_q [g(X, X)] < \infty \}$ with inner product $\lrangle{ X, Y }_{\clL^2_{q}} := \bbE_{q(x)} [g_x(X(x), Y(x))]$, and the overline means closure.
The tangent space $T_{q} \clP$ inherits an inner product from $\clL^2_{q}(\clM)$, which defines the Riemannian structure on $\clP(\clM)$.
It is consistent with the Wasserstein distance \cite{benamou2000computational}.
With this structure, the gradient of the KL divergence $\KL_p(q) := \int_{\clM} \log(q/p) \dd q$ is given explicitly (\citet{villani2008optimal}, Formula~15.2, Thm.~23.18):
\begin{align}
  \grad \KL_p (q) = \grad \log (q / p) \quad \in T_{q} \clP(\clM).
  \label{eqn:gradkl}
\end{align}
Noting that $T_{q} \clP$ is a linear subspace of the Hilbert space $\clL^2_q(\clM)$, an orthogonal projection $\pi_q: \clL^2_q(\clM) \to T_{q} \clP$ can be uniquely defined.
For any $X \in \clL^2_q(\clM)$, $\pi_q(X)$ is the unique vector in $T_{q} \clP$ such that $\divg(q X) = \divg(q \pi_q(X))$ (\citet{ambrosio2008gradient}, Lem.~8.4.2),
where $\divg$ is the divergence on $\clM$ and $\divg(q X) = \partial_i (q X^i)$ when $q$ is the density w.r.t. the Lebesgue measure of the coordinate space $\bbR^M$.
The projection can also be explained with a physical intuition.
Let $X \in \clL^2_q(\clM)$ be a vector field on $\clM$, and let its flow act on the random variable $x$ of $q$.
The transformed random variable $\varphi_t(x)$ specifies a distribution $q_t$, and a distribution curve $(q_t)_t$ is then induced by $X$.
The tangent vector of such $(q_t)_t$ at $q$ is exactly $\pi_q(X)$.

\subsubsection{Hamiltonian Flows}
\label{sec:pre-flow-ham}

The Hamiltonian flow is an abstraction of the Hamiltonian dynamics in classical mechanics \cite{marsden2013introduction}.
It is defined in association to a \emph{Poisson structure} (\citet{fernandes2014lectures}) on a manifold $\clM$, which can be expressed either as a Poisson bracket $\{\cdot,\cdot\}: \clC^{\infty}(\clM) \times \clC^{\infty}(\clM) \to \clC^{\infty}(\clM)$, 
or equivalently as a bivector field $\beta: T^* \! \clM \times T^* \! \clM \to \clC^{\infty}(\clM)$ via the relation $\beta(\ud f, \ud h) = \{f, h\}$. 
Expressed in coordinates, $\beta_x(\ud f (x), \ud h (x)) = \beta^{ij}(x) \partial_i f (x) \partial_j h (x)$, where the matrix $(\beta^{ij}(x))$ is required to be skew-symmetric and satisfy:
\begin{align}
  \beta^{il} \partial_l \beta^{jk} + \beta^{jl} \partial_l \beta^{ki} + \beta^{kl} \partial_l \beta^{ij} = 0, \forall i,j,k.
  \label{eqn:jacob}
\end{align}
The \emph{Hamiltonian vector field} of a smooth function $f$ on $\clM$ is defined as $X_f(\cdot) := \{\cdot, f\}$, with coordinate expression: 
\begin{align}
  X_f(x) = \beta^{ij}(x) \partial_j f(x) \partial_i \quad \in T_x \clM.
  \label{eqn:hamfl}
\end{align}
A Hamiltonian flow $\{(\varphi_t(x))_t\}$ is then determined by $X_f$.
Its key property is that it conserves $f$: $f(\varphi_t(x))$ is constant w.r.t. $t$. 
The Hamiltonian flow may be more widely known on a symplectic manifold or more particularly a cotangent bundle (\eg, \citet{da2001lectures}; \citet{marsden2013introduction}), but these cases are not general enough for our purpose (\eg, they require $\clM$ to be even-dimensional).

On $\clP(\clM)$, a Poisson structure can be induced by the one $\{\cdot, \cdot\}_{\clM}$ of $\clM$.
Consider linear functions on $\clP(\clM)$ in the form $F_f: q \mapsto \bbE_{q} [f]$ for $f \in \clCC(\clM)$.
A Poisson bracket for these linear functions can be defined as (\eg, \citet{lott2008some}, Sec.~6; \citet{gangbo2010differential}, Sec.~7.2):
\begin{align}
  \{F_f, F_h\}_{\clP} := F_{\{f, h\}_{\clM}}.
  \label{eqn:wpois}
\end{align}
This bracket can be extended for any smooth function $F$ by its \emph{linearization} at $q$, which is a linear function $F_f$ such that $\grad F_f(q) \! = \! \grad F(q)$.
The extended bracket is then given by $\{F, H\}_{\clP}(q) \! := \! \{F_f, F_h\}_{\clP}(q)$ (\citet{gangbo2010differential}, Rem.~7.8), where $F_f$, $F_h$ are the linearizations at $q$ of functions $F$, $H$.
The Hamiltonian vector field of $F$ is then identified as (\citet{gangbo2010differential}, Sec.~7.2):
\begin{align}
  \clX_F(q) = \clX_{F_f}(q) = \pi_q(X_f) \quad \in T_{q} \clP(\clM).
  \label{eqn:wham}
\end{align}

\vspace{-4pt}
On the same topic, \citet{ambrosio2008hamiltonian} study the existence and simulation of the Hamiltonian flow on $\clP(\clM)$ for $\clM$ as a symplectic Euclidean space, and verify the conservation of Hamiltonian under certain conditions.
\citet{gangbo2010differential} investigate the Poisson structure on the algebraic dual $(\clCC(\clM))^*$, a superset of $\clP(\clM)$, and find that the canonical Poisson structure induced by the Lie structure of $\clCC(\clM)$ coincides with \eqref{eqn:wpois}.
Their consideration is also for symplectic Euclidean $\clM$, but the procedures and conclusions can be directly adapted to Riemannian Poisson manifolds.
\citet{lott2008some} considers the Poisson structure \eqref{eqn:wpois} on the space of smooth distributions on a Poisson manifold $\clM$, and find that it is the restriction of the Poisson structure of $(\clCC(\clM))^*$ by \citet{gangbo2010differential}.

\section{Understanding MCMC Dynamics as Flows on the Wasserstein Space $\clP(\clM)$}
\label{sec:flow}

This part presents our main discovery that connects MCMC dynamics and flows on the Wasserstein space $\clP(\clM)$.
We first work on the two concepts and introduce novel concepts for preparation, then propose the unified framework and analyze existing MCMC instances under the framework.

\subsection{Technical Development}
\label{sec:flow-prep}

We excavate into MCMC and Wasserstein flows and introduce novel concepts in preparation for the framework.

\textbf{On the MCMC side}~~
Noting that flows on $\clP(\clM)$ are deterministic while MCMCs involve stochastic diffusion, we first reformulate MCMC dynamics as an equivalent deterministic one for unification.
Here we say two dynamics are \emph{equivalent} if they produce the same distribution curve.
\begin{lemma}[Equivalent deterministic MCMC dynamics]
  \label{lem:detmcmc}
  The MCMC dynamics \eqref{eqn:recipe} with symmetric diffusion matrix $D$ is equivalent to the deterministic dynamics in $\bbR^M$:
  \vspace{-10pt}
  \begin{equation}\begin{gathered}
	\ud x = W_t(x) \dd t, \\
	(W_t)^i = D^{ij} \partial_j \log (p / q_t) + Q^{ij} \partial_j \log p + \partial_j Q^{ij},
	\label{eqn:detmcmc}
  \end{gathered}\end{equation}
  where $q_t$ is the distribution density of $x$ at time $t$.
\end{lemma}
Proof is provided in Appendix~A.1.
For any $q \in \clP(\bbR^M)$, the projected vector field $\pi_{q}(W)$ can be treated as a tangent vector at $q$, so $W$ defines a vector field on $\clP(\bbR^M)$.
In this way, we give a first view of an MCMC dynamics as a Wasserstein flow. 
An equivalent flow with a richer structure will be given in Theorem~\ref{thm:equiv}. 

This expression also helps understanding Barbour's generator $\clA$ \cite{barbour1990stein} of an MCMC dynamics, which can be used in Stein's method~\cite{stein1972bound} of constructing distribution metrics.
For instance the standard Langevin dynamics induces the Stein's operator, and it in turn produces a metric called the Stein discrepancy~\cite{gorham2015measuring}, which inspires SVGD,
and \citet{liu2018riemannian} consider the Riemannian counterparts.
The Barbour's generator maps a function $f \in \clCC(\bbR^M)$ to another $(\clA f) (x) := \frac{\ud}{\ud t} \bbE_{q_t} [f] \big|_{t=0}$,
where $(q_t)_t$ obeys initial condition $q_0 = \delta_x$ (Dirac measure).
In terms of the linear function $F_f$ on $\clP(\bbR^M)$, we recognize $(\clA f) (x) = \frac{\ud}{\ud t} F_f (q_t) \big|_{t=0} = \lrangle{ \grad F_f, \pi_{q_0}(W_0) }_{T_{q_0}\clP}$ as the \emph{directional derivative} of $F_f$ along $(q_t)_t$ at $q_0$.
This knowledge gives the expression:
\begin{align}
  \clA f = \frac{1}{p} \partial_j \left[ p \left( D^{ij} + Q^{ij} \right) (\partial_i f) \right],
  \label{eqn:barbour}
\end{align}
which meets existing results (\eg, \citet{gorham2016measuring}, Thm.~2).
Details are provided in Appendix~A.2.

\textbf{On the Wasserstein flow side}~~
We deepen the knowledge on flows on $\clP(\clM)$ with a Riemannian and Poisson structure of $\clM$.\footnote{
  We do not consider the compatibility of the Riemannian and Poisson structure so it is different from 
  a K\"{a}hler manifold.
}
The gradient of $\KL_p$ is given by \eqref{eqn:gradkl}, but its Hamiltonian vector field is not directly available due to its non-linearity.
We first develop an explicit expression for it.
\begin{lemma}[Hamiltonian vector field of KL on $\clP(\clM)$]
  \label{lem:klham}
  Let $\beta$ be the bivector field form of a Poisson structure on $\clM$ and $\clP(\clM)$ endowed with the induced Poisson structure described in Section~\ref{sec:pre-flow-ham}.
  Then the Hamiltonian vector field of $\KL_p$ on $\clP(\clM)$ is:
  \begin{align}
	\clX_{\KL_p} (q) = \pi_q( X_{\log (q/p)} ) = \pi_q( \beta^{ij} \partial_j \log (q/p) \partial_i ).
	\label{eqn:klham}
  \end{align}
\end{lemma}
Proof is provided in Appendix~A.3.
Note that the projection $\pi_q$ does not make much difference recalling $X$ and $\pi_q(X)$ produce the same distribution curve through $q$. 

For a wider coverage of our framework on MCMC dynamics, we introduce a novel concept called \emph{fiber-Riemannian manifold} and develop associated objects.
This notion generalizes Riemannian manifold, such that the non-degenerate requirement of the Riemannian structure is relaxed.
\begin{definition}[Fiber-Riemannian manifold]
  We say that a manifold $\clM$ is a \emph{fiber-Riemannian manifold} if it is a fiber bundle and there is a Riemannian structure on each \emph{fiber}.
\end{definition}
\begin{wrapfigure}{r}{.24\textwidth}
  \centering
  \vspace{-6pt}
  \includegraphics[width=.26\textwidth]{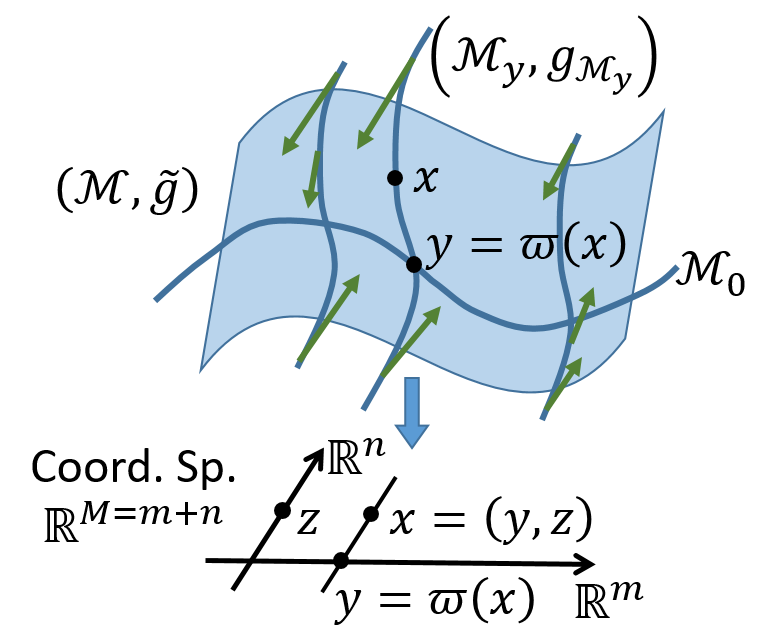}
  \vspace{-20pt}
  \caption{Illustration of a fiber-Riemannian manifold $(\clM, \tgg)$ ($m=n=1$) and a fiber-gradient shown in green arrows.}
  \vspace{-10pt}
  \label{fig:fiber}
\end{wrapfigure}
See Fig.~\ref{fig:fiber} for illustration.
Roughly, $\clM$ (of dimension $M = m + n$) is a fiber bundle if there are two smooth manifolds $\clM_0$ (of dimension $m$) and $\clF$ (of dimension $n$) and a surjective projection $\varpi: \clM \to \clM_0$ such that $\varpi$ is locally equivalent to the projection on the product space $\clM_0 \times \clF \to \clM_0$ (\eg, \citet{nicolaescu2007lectures}, Def.~2.1.21).
The space $\clM_0$ is called the base space, and $\clF$ the common fiber.
The \emph{fiber} through $x \in \clM$ is defined as the submanifold $\clM_{\varpi(x)} := \varpi^{-1}(\varpi(x))$, which is diffeomorphic to $\clF$.
Fiber bundle generalizes the concept of the product space to allow different structures among different fibers.
The coordinate of $\clM$ can be decomposed under this structure: $x = (y, z)$ where $y \in \bbR^m$ is the coordinate of $\clM_0$ and $z \in \bbR^n$ of $\clM_{\varpi(x)}$.
Coordinates of points on a fiber share the same $y$ part.
We allow $m$ or $n$ to be zero.

According to our definition, a fiber-Riemannian manifold furnish each fiber $\clM_y$ with a Riemannian structure $g_{\clM_y}$, whose coordinate expression is $\big( (g_{\clM_y}\!)_{ab} \big)$ (indices $a, b$ for $z$ run from $1$ to $n$).
By restricting a function $f \in \clC^{\infty} (\clM)$ on a fiber $\clM_y$, the structure defines a gradient on the fiber: $\grad_{\clM_y} f(y, z) = (g_{\clM_y}\!)^{ab}(z) \, \partial_{z^b} f(y, z) \, \partial_{z^a}$.
Taking the union over all fibers, we have a vector field on the entire manifold $\clM$, which we call the \emph{fiber-gradient} of $f$: $\big( (\gradf f )^i (x) \big) := \big( 0_m, (g_{\clM_{\varpi(x)}}\!)^{ab}(z) \, \partial_{z^b} f(\varpi(x), z) \big)$.
To express it in a similar way as the gradient, we further define the \emph{fiber-Riemannian structure} $\tgg$ as:
\begin{align}
  \big( \tgg^{ij} (x) \big)_{M \times M} := \begin{pmatrix}
	0_{m \times m} & 0_{m \times n} \\
	0_{n \times m} & \big( (g_{\clM_{\varpi(x)}}\!)^{ab} (z) \big)_{n \times n}
  \end{pmatrix},
  \!
  \label{eqn:fibriem}
\end{align}
and the fiber-gradient can be expressed as $\gradf f = \tgg^{ij} \partial_j \! f \partial_i$.
Note that $\gradf f (x)$ is tangent to the fiber $\clM_{\varpi(x)}$ and its flow moves points within each fiber. 
It is not a Riemannian manifold for $m \ge 1$ since $(\tgg^{ij})$ is singular.

Now we turn to the Wasserstein space.
As the fiber structure of $\clP(\clM)$ is hard to find, we consider the space $\tclP(\clM) := \Set{q(\cdot|y) \in \clP(\clM_y)}{y \in \clM_0}$.
With projection $q(\cdot|y) \mapsto y$, it is locally equivalent to $\clM_0 \times \clP(\clM_y)$.
Each of its fiber $\clP(\clM_y)$ has a Riemannian structure induced by that of $\clM_y$ (Section~\ref{sec:pre-flow-grad}), so it is a fiber-Riemannian manifold.
On fiber $\clP(\clM_y)$, according to \eqref{eqn:gradkl}, we have $\grad \KL_{p(\cdot|y)} \! \big( q(\cdot|y) \big) (z) = (g_{\clM_y}\!)^{ab}(z) \partial_{z^b} \! \log \! \frac{q(z|y)}{p(z|y)} \partial_{z^a} = (g_{\clM_y}\!)^{ab}(z) \partial_{z^b} \! \log \frac{q(y,z)}{p(y,z)} \partial_{z^a}$ as a vector field on $\clM_y$.
Taking the union over all fibers, we have the fiber-gradient of $\KL_p$ on $\tclP(\clM)$ as a vector field on $\clM$:
\begin{align}
  \gradf \KL_p (q) (x) = \tgg^{ij}(x) \, \partial_j \log \big( q(x) / p(x) \big) \, \partial_i.
  \label{eqn:fibgrad}
\end{align}
After projected by $\pi_q$, $\gradf \KL_p (q)$ is a tangent vector on the Wasserstein space $\clP(\clM)$.
Note that $\clP(\clM)$ is locally equivalent to $\clP(\clM_0) \times \tclP(\clM)$ thus not a fiber-Riemannian manifold in this way, so it is hard to develop the fiber-gradient directly on $\clP(\clM)$.

\subsection{The Unified Framework}
\label{sec:flow-frame}

We introduce a \emph{regularity} assumption on MCMC dynamics that our unified framework considers.
It is satisfied by almost all existing MCMCs and its relaxation will be discussed at the end of this section.
\begin{assumption}[Regular MCMC dynamics]
  \label{asm:reg}
  We call an MCMC dynamics regular if its corresponding matrices $(D, Q)$ in formulation~\origeqref{eqn:recipe} additionally satisfies:
  \textbf{(a)} the diffusion matrix $D = C$ or $D = 0$ or $D = \begin{pmatrix} 0 & 0 \\ 0 & C \end{pmatrix}$, where $C(x)$ is symmetric positive definite everywhere;
  \textbf{(b)} the curl matrix $Q(x)$ satisfies \eqref{eqn:jacob} everywhere.
\end{assumption}

\begin{figure}[t]
  \centering
  \includegraphics[width=.45\textwidth]{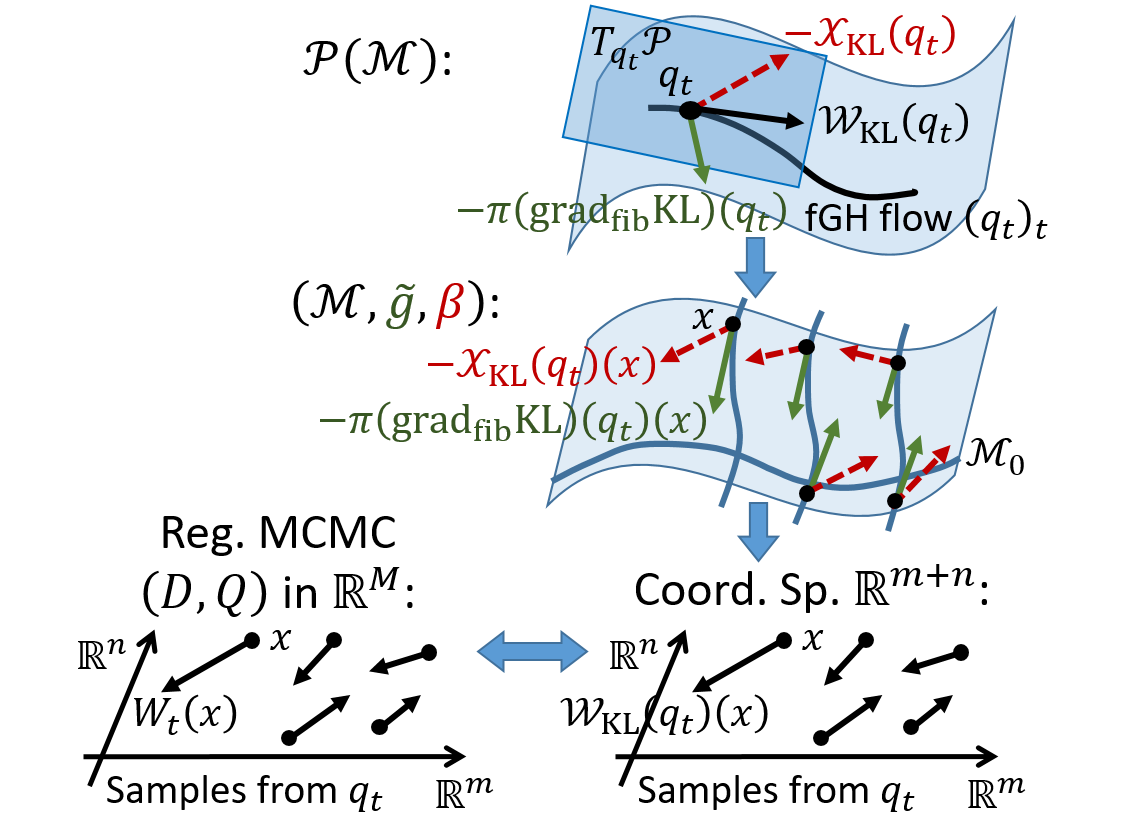}
  \vspace{-12pt}
  \caption{Illustration of our unified framework (Theorem~\ref{thm:equiv}): a regular MCMC dynamics is equivalent to the fGH flow $\clW_{\KL_p}$ on the Wasserstein space $\clP(\clM)$ of an fRP manifold $\clM$.
	The projected fiber-gradient (green solid arrows) and Hamiltonian vector field (red dashed arrows) at $q_t$ on $\clM$ are plotted.
  }
  \vspace{-8pt}
  \label{fig:frame}
\end{figure}
Now we formally state our unified framework, with an illustration provided in Fig.~\ref{fig:frame}.
\begin{theorem}[\textbf{Unified framework}: equivalence between regular MCMC dynamics and fGH flows on $\clP(\clM)$]
  \label{thm:equiv}
  We call $(\clM, \tgg, \beta)$ a fiber-Riemannian Poisson (fRP) manifold, and define the fiber-gradient Hamiltonian (fGH) flow on $\clP(\clM)$ as the flow induced by the vector field:
  \begin{equation}\begin{split}
	\clW_{\KL_p} :=& - \pi( \gradf \KL_p ) - \clX_{\KL_p}, \\
	\clW_{\KL_p} (q) =& \pi_q \big(\, (\tgg^{ij} + \beta^{ij}) \partial_j \log (p/q) \partial_i \,\big).
	\label{eqn:fgh}
  \end{split}\end{equation}
  Then:
  \textbf{(a)} Any regular MCMC dynamics on $\bbR^M$ targeting $p$ is equivalent to the fGH flow $\clW_{\KL_p}$ on $\clP(\clM)$ for a certain fRP manifold $\clM$;
  \textbf{(b)} Conversely, for any fRP manifold $\clM$, the fGH flow $\clW_{\KL_p}$ on $\clP(\clM)$ is equivalent to a regular MCMC dynamics targeting $p$ in the coordinate space of $\clM$;
  \textbf{(c)} More precisely, in both cases, the coordinate expressions of the fiber-Riemannian structure $\tgg$ and Poisson structure $\beta$ of $\clM$ coincide respectively with the diffusion matrix $D$ and the curl matrix $Q$ of the regular MCMC dynamics.
\end{theorem}
The idea of proof is to show $\pi_q (W) = \clW_{\KL_p} (q)$ ($W$ defined in Lemma~\ref{lem:detmcmc}) at any $q \in \clP(\clM)$ so that the two vector fields produce the same evolution rule of distribution.
Proof details are presented in Appendix~A.4.

This formulation unifies regular MCMC dynamics and flows on the Wasserstein space, and provides a direct explanation on the behavior of general MCMC dynamics.
The fundamental requirement on MCMCs that the target distribution $p$ is kept stationary, turns obvious in our framework: $\clW_{\KL_p}(p) = 0$.
The Hamiltonian flow $-\clX_{\KL_p}$ conserves $\KL_p$ (difference to $p$) while encourages efficient exploration in the sample space that helps faster convergence and lower autocorrelation \cite{betancourt2017geometric}.
The fiber-gradient flow $-\gradf \KL_p$ minimizes $\KL_{p(\cdot|y)}$ on each fiber $\clM_y$, driving $q_t(\cdot|y)$ to $p(\cdot|y)$ and enforcing convergence.
Specification of this general behavior is discussed below.

\subsection{Existing MCMCs under the Unified Framework}
\label{sec:flow-ins}

Now we make detailed analysis on existing MCMC methods under our unified framework.
Depending on the diffusion matrix $D$, they can be categorized into three types.
Each type has a particular fiber structure of the corresponding fRP manifold, thus a particular behavior of the dynamics.

\textbf{Type 1:} $D$ is non-singular ($m=0$ in \eqref{eqn:fibriem}). \\
In this case, the corresponding $\clM_0$ degenerates and $\clM$ itself is the unique fiber, so $\clM$ is a Riemannian manifold with structure $(g_{ij}) = D^{-1}$.
The fiber-gradient flow on $\tclP(\clM)$ becomes the gradient flow on $\clP(\clM)$ so:
\begin{align}
  \clW_{\KL_p} = -\pi( \grad \KL_p ) - \clX_{\KL_p},
\end{align}
which indicates the convergence of the dynamics: the Hamiltonian flow $-\clX_{\KL_p}$ conserves $\KL_p$ while the gradient flow $-\grad \KL_p$ minimizes $\KL_p$ on $\clP(\clM)$ steepest, so they jointly minimize $\KL_p$ monotonically, leading to the unique minimizer $p$.
This meets the conclusion in \citet{ma2015complete}.

The Langevin dynamics (LD) \cite{roberts1996exponential}, used in both full-batch~\cite{roberts2002langevin} and stochastic gradient (SG) simulation~\cite{welling2011bayesian}, falls into this class.
Its curl matrix $Q = 0$ makes its fGH flow comprise purely the gradient flow, allowing a rich study on its behavior \cite{durmus2016high, cheng2017convergence, wibisono2018sampling, bernton2018langevin, durmus2018analysis}.
Its Riemannian version~\cite{girolami2011riemann} chooses $D$ as the inverse Fisher metric so that $\clM$ is the distribution manifold in information geometry~\cite{amari2016information}.
\citet{patterson2013stochastic} further explore the simulation with SG.

\textbf{Type 2:} $D = 0$ ($n=0$ in \eqref{eqn:fibriem}). \\
In this case, $\clM_0 = \clM$ and fibers degenerate.
The fGH flow $\clW_{\KL_p}$ comprises purely the Hamiltonian flow $-\clX_{\KL_p}$, which conserves $\KL_p$ and helps distant exploration.
We note that under this case, the decrease of $\KL_p$ is not guaranteed, so care must be taken in simulation.
Particularly, this type of dynamics cannot be simulated with parallel chains unless samples initially distribute as $p$, so they are not suitable for ParVI simulation.
The lack of a stabilizing force in the dynamics also explains their vulnerability in face of SG, where the noisy perturbation is uncontrolled.
This generalizes the discussion on HMC by \citet{chen2014stochastic} and \citet{betancourt2015fundamental} to dynamics of this type.

The Hamiltonian dynamics (\eg, \citet{marsden2013introduction}, Chap.~2) that HMC simulates is a representative of this kind.
To sample from a distribution $p(\theta)$ on manifold $\clS$ of dimension $\ell$, variable $\theta$ is augmented $x=(\theta,r)$ with a vector $r \in \bbR^{\ell}$ called momentum.
In our framework, this is to take $\clM$ as the cotangent bundle $T^* \! \clS$, 
whose canonical Poisson structure corresponds to $Q = (\beta^{ij}) = \begin{pmatrix} 0 & -I_{\ell} \\ I_{\ell} & 0 \end{pmatrix}$.
A conditional distribution $p(r|\theta)$ 
is chosen for an augmented target distribution $p(x) = p(\theta) p(r|\theta)$.
HMC produces more effective samples than LD with the help of the Hamiltonian flow \cite{betancourt2017geometric}.
As we mentioned, the dynamics of HMC cannot guarantee convergence, so it relies on the \emph{ergodicity} of its simulation for convergence \cite{livingstone2016geometric, betancourt2017conceptual}.
It is simulated in a deliberated way: the second-order symplectic leap-frog integrator is employed, and $r$ is successively redrew from $p(r|\theta)$. 

HMC considers Euclidean $\clS$ and chooses Gaussian $p(r|\theta) = \clN(0, \Sigma)$, while \citet{zhang2016towards} take $p(r|\theta)$ as the monomial Gamma distribution.
On Riemannian $(\clS, g)$, $p(r|\theta)$ is chosen as $\clN \big( 0, (g_{ij}(\theta)) \big)$, \ie, the standard Gaussian in the cotangent space $T^*_{\theta} \clS$ \cite{girolami2011riemann}.
\citet{byrne2013geodesic} simulate the dynamics for manifolds with no global coordinates, and \citet{lan2015markov} take the Lagrangian form for better simulation, which uses velocity (tangent vector) in place of momentum (covector).

\textbf{Type 3:} $D \neq 0$ and $D$ is singular ($m, n \ge 1$ in \eqref{eqn:fibriem}). \\
In this case, both $\clM_0$ and fibers are non-degenerate. 
The fiber-gradient flow stabilizes the dynamics only in each fiber $\clM_y$, but this is enough for most SG-MCMCs since SG appears only in the fibers.

SGHMC \cite{chen2014stochastic} is the first instance of this type.
Similar to the Hamiltonian dynamics, it takes $\clM = T^* \! \clS$ and shares the same $Q$, but its $D_{2\ell \times 2\ell}$ is in the form of Assumption~\ref{asm:reg}(a) with a constant $C_{\ell \times \ell}$, whose inverse $C^{-1}$ defines a Riemannian structure in every fiber $\clM_y$. 
Viewed in our framework, this makes the fiber bundle structure of $\clM$ coincides with that of $T^* \! \clS$: $\clM_0 = \clS$, $\clM_y = T^*_{\theta} \clS$, and $x = (y,z) = (\theta,r)$.
Using Lemma~\ref{lem:detmcmc}, with a specified $p(r|\theta)$, 
we derive its equivalent deterministic dynamics:
\begin{align}
  \!\!
  \begin{cases}
	\! \frac{\ud \theta}{\ud t} = - \nabla_r \log p(r|\theta), \\
	\! \frac{\ud r}{\ud t} = \nabla_{\theta} \! \log p(\theta) \!+\! \nabla_{\theta} \! \log p(r|\theta) \!+ C \nabla_r \! \log \frac{p(r|\theta)}{q(r|\theta)}.
  \end{cases}
  \!\!\!\!\!\!\!\!
  \label{eqn:sghmc-det}
\end{align}
We note that it adds the dynamics $\frac{\ud r}{\ud t} = C \nabla_r \log \frac{p(r|\theta)}{q(r|\theta)}$ to the Hamiltonian dynamics.
This added dynamics is essentially the fiber-gradient flow $- (\gradf \KL_p) (q)$ on $\clP(\clM)$ (\eqref{eqn:fibgrad}), or the gradient flow $- ( \grad \KL_{p(\cdot|\theta)} ) (q(\cdot|\theta))$ on fiber $T^*_{\theta} \clS$, which pushes $q(\cdot|\theta)$ towards $p(\cdot|\theta)$.
In presence of SG, the dynamics for $\theta \in \clS$ is unaffected, but for $r \in T^*_{\theta} \clS$ in each fiber, a fluctuation is introduced due to the noisy estimate of $\nabla_{\theta} \log p(\theta)$, which will mislead $q(\cdot|\theta)$.
The fiber-gradient compensates this by guiding $q(\cdot|\theta)$ to the correct target, making the dynamics robust to SG.

Another famous example of this kind is the SG Nos\'{e}-Hoover thermostats (SGNHT)~\cite{ding2014bayesian}.
It further augments $(\theta,r)$ with the thermostats $\xi \in \bbR$ to better balance the SG noise.
In terms of our framework, the thermostats $\xi$ augments $\clM_0$, and the fiber is the same as SGHMC.

Both SGHMC and SGNHT choose $p(r|\theta) = \clN(0, \Sigma^{-1})$, while SG monomial Gamma thermostats (SGMGT) \cite{zhang2017stochastic} uses monomial Gamma, and \citet{lu2016relativistic} choose $p(r|\theta)$ according to a relativistic energy function to adapt the scale in each dimension. 
Riemannian extensions of SGHMC and SGNHT on $(\clS, g)$ are explored by \citet{ma2015complete} and \citet{liu2016stochastic}. 
Viewed in our framework, they induce a Riemannian structure $\big( \! \sqrt{\! (g^{ij}(\theta))} \trs \! C^{-1} \! \sqrt{\! (g^{ij}(\theta))} \big)_{\ell \times \ell}$ in each fiber $\clM_y = T^*_{\theta} \clS$.

\textbf{Discussions}~~
Due to the linearity of the equivalent systems~\origeqref{eqn:recipe},~\origeqref{eqn:detmcmc},~\origeqref{eqn:fgh} w.r.t. $D$, $Q$ or $(\tgg^{ij})$, $(\beta^{ij})$, MCMC dynamics can be combined.
From the analysis above, SGHMC can be seen as the combination of the Hamiltonian dynamics on the cotangent bundle $T^* \! \clS$ and the LD in each fiber (cotangent space $T^*_{\theta} \clS$).
As another example, \citet{zhang2017stochastic} combine SGMGT of Type 3 with LD of Type 1, creating a Type 1 method that decreases $\KL_p$ on the entire manifold instead of each fiber.
This improves the convergence, which meets their empirical observation.

Assumption~\ref{asm:reg}(a) is satisfied by all the mentioned MCMC dynamics, and Assumption~\ref{asm:reg}(b) is also satisfied by all except SGNHT related dynamics. 
On this exception, we note from the derivation of Theorem~\ref{thm:equiv} that, Assumption~\ref{asm:reg}(b) is only required for $\clM$ thus $\clP(\clM)$ to be a Poisson manifold, but is not used in the deduction afterwards.
Definition of a Hamiltonian vector field and its key property could also be established without the assumption, so it is possible to extend the framework under a more general mathematical concept that relaxes Assumption~\ref{asm:reg}(b).
Assumption~\ref{asm:reg}(a) could also be hopefully relaxed by an invertible transformation from any positive semi-definite $D$ into the required form, effectively converting the dynamics into an equivalent regular one.
We leave further investigations as future work.

\section{Simulation as ParVIs}
\label{sec:parvi}

The unified framework (Theorem~\ref{thm:equiv}) recognizes an MCMC dynamics as an fGH flow on the Wasserstein space $\clP(\clM)$ of an fRP manifold $\clM$, expressed in \eqref{eqn:fgh} explicitly.
Lemma~\ref{lem:detmcmc} gives another equivalent dynamics that leads to the same flow on $\clP(\clM)$.
These findings enable us to simulate these flow-based dynamics for an MCMC method, using existing finite-particle flow simulation methods in the ParVI field.
This hybrid of ParVI and MCMC largely extends the ParVI family with various dynamics, and also gives advantages like particle-efficiency to MCMCs.

We select the SGHMC dynamics as an example and develop its particle-based simulations.
With $p(r|\theta) = \clN(0, \Sigma)$ for a constant $\Sigma$, $r$ and $\theta$ become independent, and \eqref{eqn:sghmc-det} from Lemma~\ref{lem:detmcmc} becomes:
\begin{align}
  \begin{cases}
	\frac{\ud \theta}{\ud t} = \Sigma^{-1} r, \\
	\frac{\ud r}{\ud t} = \nabla_{\theta} \log p(\theta) - C \Sigma^{-1} r - C \nabla_r \log q(r).
  \end{cases}
  \label{eqn:psghmcd}
\end{align}
From the other equivalent dynamics given by the framework (Theorem~\ref{thm:equiv}), the fGH flow (\eqref{eqn:fgh}) for SGHMC is:
\begin{align}
  \hspace{-10pt}
  \begin{cases}
	\! \frac{\ud \theta}{\ud t} = \Sigma^{-1} r + \nabla_r \log q(r), \\
	\! \frac{\ud r}{\ud t} = \nabla_{\theta}\! \log p(\theta) \!-\! C \Sigma^{-1} r \!-\!  C \nabla_r\! \log q(r) \!-\! \nabla_{\theta}\! \log q(\theta).
  \end{cases}
  \hspace{-37pt} \label{eqn:psghmcf}
\end{align}
The key problem in simulating these flow-based dynamics with finite particles is that the density $q$ is unknown.
\citet{liu2019understanding_a} give a summary on the solutions in the ParVI field, and find that they are all based on a smoothing treatment, in a certain formulation of either smoothing the density or smoothing functions.
Here we adopt the Blob method~\cite{chen2018unified} that smooths the density.
With a set of particles $\{ r^{(i)} \}_i$ of $q(r)$, Blob makes the following approximation with a kernel function $K_r$ for $r$:
\begin{align}
  \!\! - \nabla_{\! r} \! \log q( r^{(i)} \!) \! \approx - \frac{\sum_k \!\! \nabla_{\! r^{(i)}} \! K_r^{(i,k)}}{\sum_j \! K_r^{(i,j)}} \!-\! \sum_k \! \frac{\nabla_{\! r^{(i)}} \! K_r^{(i,k)}}{\sum_j \! K_r^{(j,k)}},
  \label{eqn:blob}
\end{align}
where $K_r^{(i,j)} := K_r(r^{(i)}, r^{(j)})$.
Approximation for $- \nabla_{\theta} \log q(\theta)$ can be established in a similar way.
The vanilla SGHMC simulates dynamics~\origeqref{eqn:psghmcd} with $- C \nabla_{r} \log q(r) \dd t$ replaced by $\clN (0, 2 C \dd t)$, but dynamics~\origeqref{eqn:psghmcf} cannot be simulated in a similar stochastic way.
More discussions are provided in Appendix~B.

We call the ParVI simulations of the two dynamics as pSGHMC-det (\eqref{eqn:psghmcd}) and pSGHMC-fGH (\eqref{eqn:psghmcf}), respectively (``p'' for ``particle''). 
Compared to the vanilla SGHMC, the proposed methods could converge faster and be more particle-efficient with deterministic update and explicit repulsive interaction (\eqref{eqn:blob}).
On the other hand, SGHMC could make a more efficient exploration and converges faster than LD, 
so our methods could speed up over Blob.
One may note that pSGHMC-det resembles a direct application of stochastic gradient descent with momentum \cite{sutskever2013importance} to Blob.
We stress that this application is inappropriate since Blob minimizes $\KL_p$ on the infinite-dimensional manifold $\clP(\clM)$ instead of a function on $\clM$.
Moreover, the two methods can be nourished with advanced techniques in the ParVI field.
This includes the HE bandwidth selection method and acceleration frameworks by \citet{liu2019understanding_a}, and other approximations to $-\nabla \log q$ like SVGD and GFSD/GFSF \cite{liu2019understanding_a}.

\vspace{-1pt}
\section{Experiments}
\label{sec:exp}

Detailed experimental settings are provided in Appendix~C, and codes are available at \url{https://github.com/chang-ml-thu/FGH-flow}.

\vspace{-2pt}
\subsection{Synthetic Experiment}
\vspace{-1pt}

\begin{figure}[t]
  \centering
  \includegraphics[width=.47\textwidth]{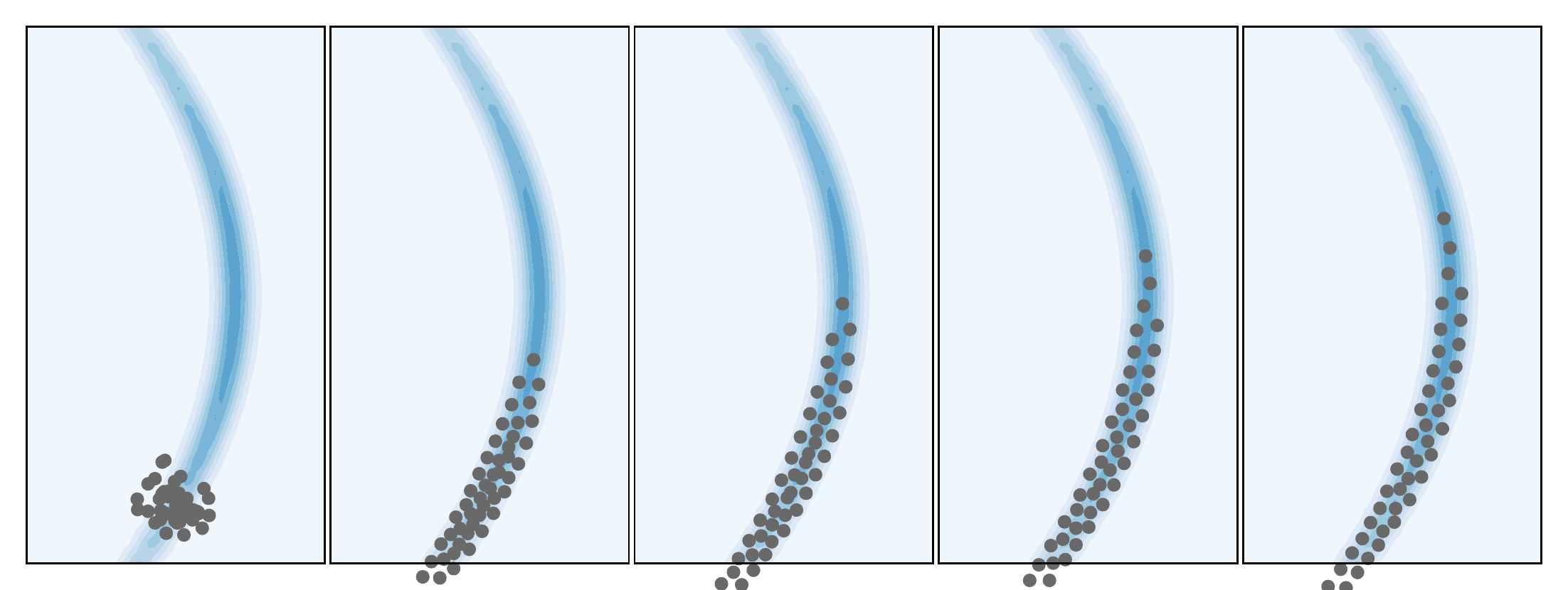} \hspace{-53pt}
  \includegraphics[width=.105\textwidth]{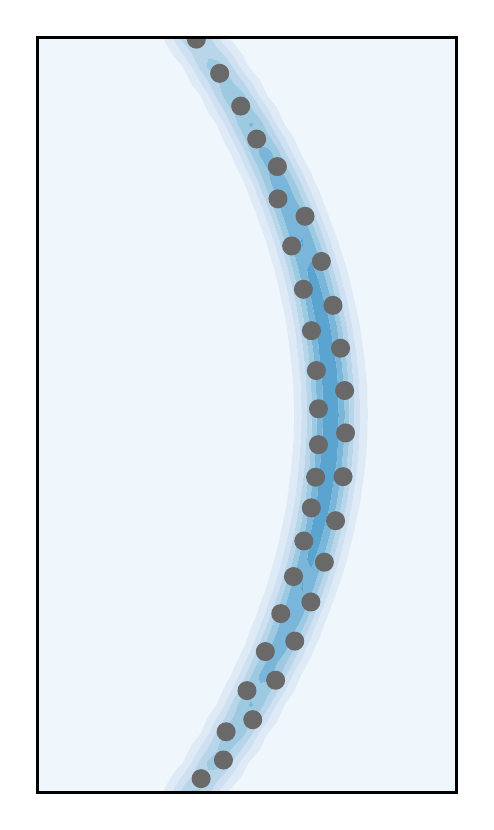}
  \\[-4.8pt]
  \includegraphics[width=.47\textwidth]{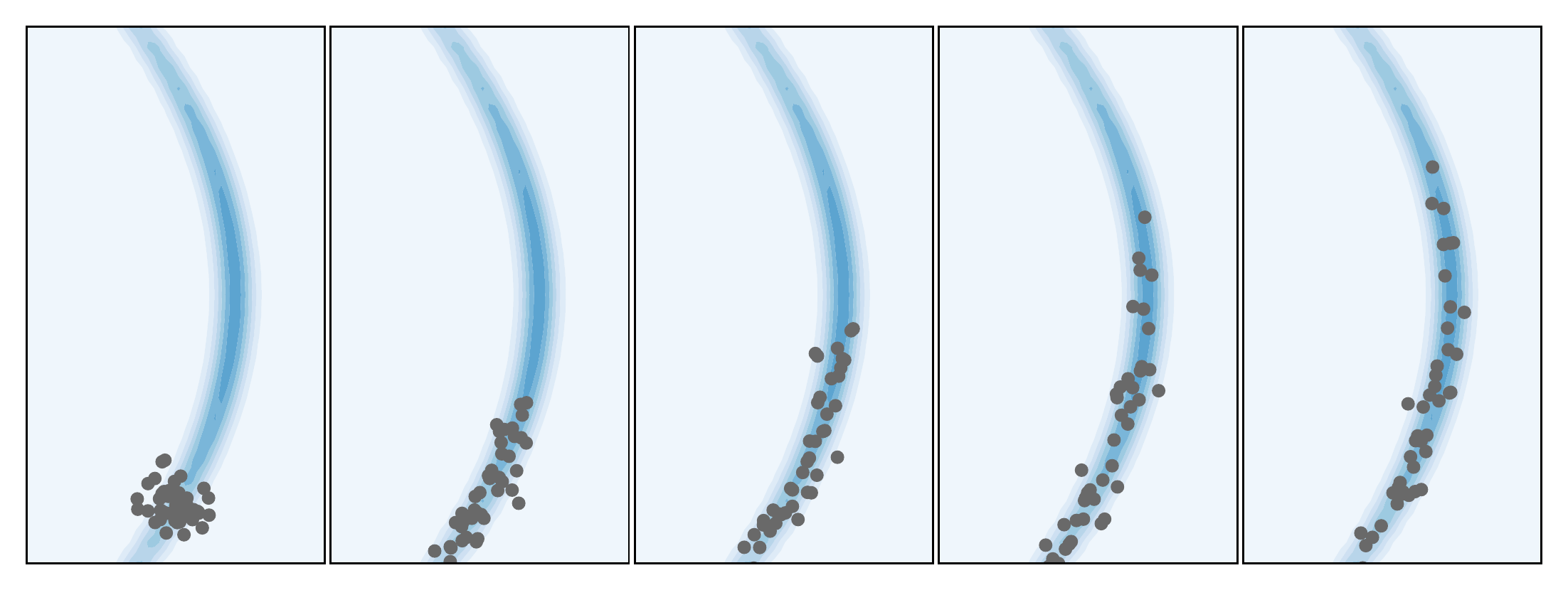} \hspace{-53pt}
  \includegraphics[width=.105\textwidth]{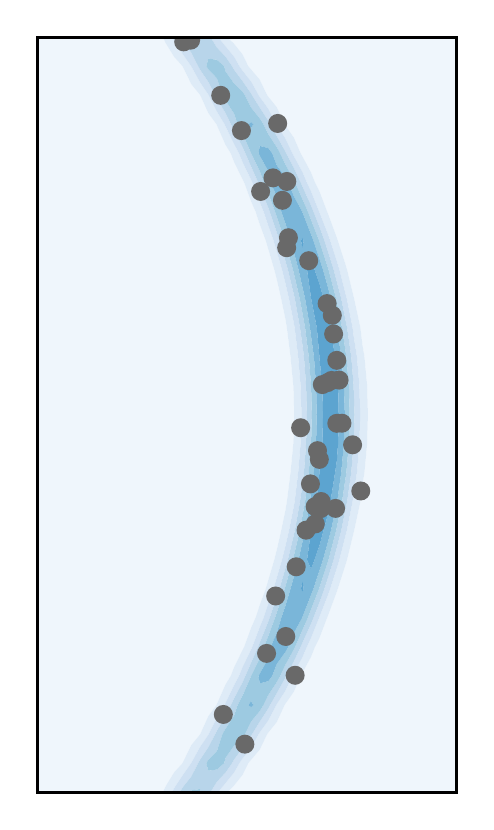}
  \\[-4.8pt]
  \includegraphics[width=.47\textwidth]{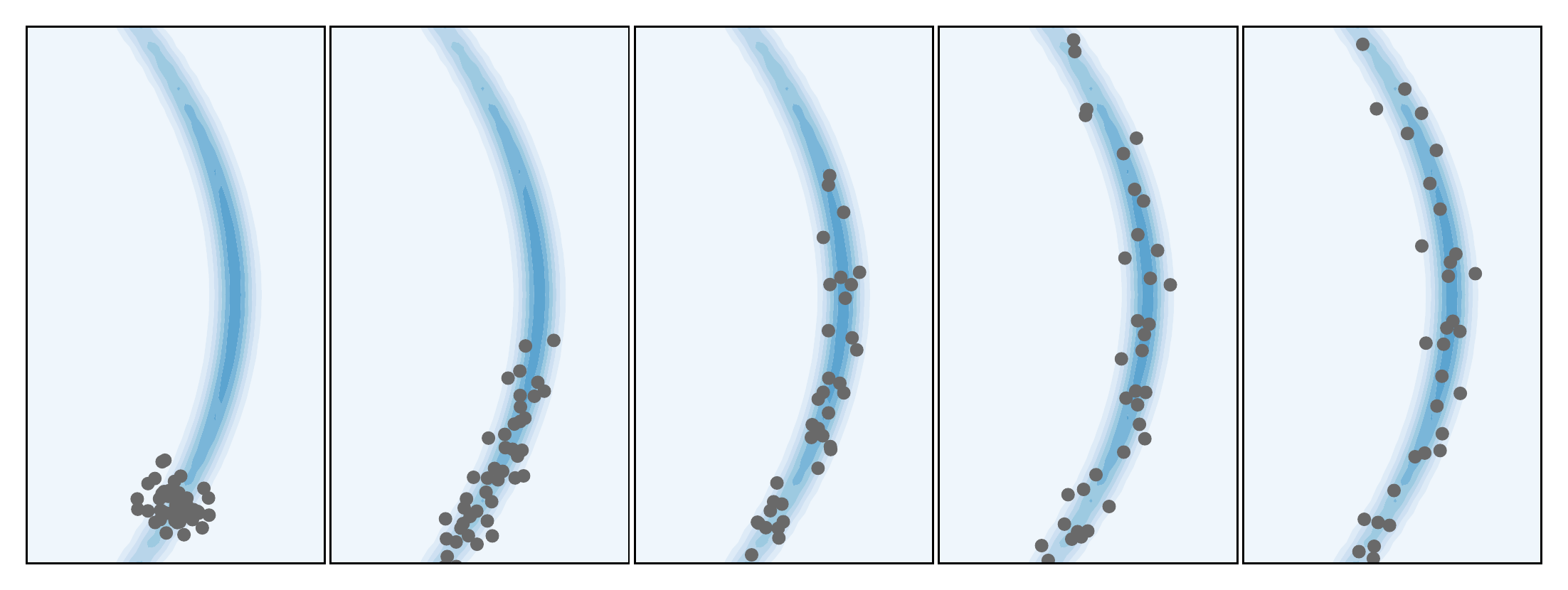} \hspace{-53pt}
  \includegraphics[width=.105\textwidth]{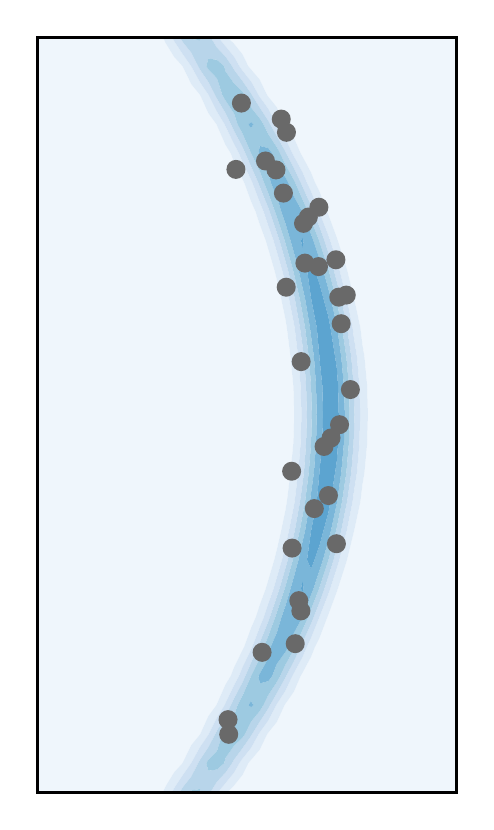}
  \\[-4.8pt]
  \includegraphics[width=.47\textwidth]{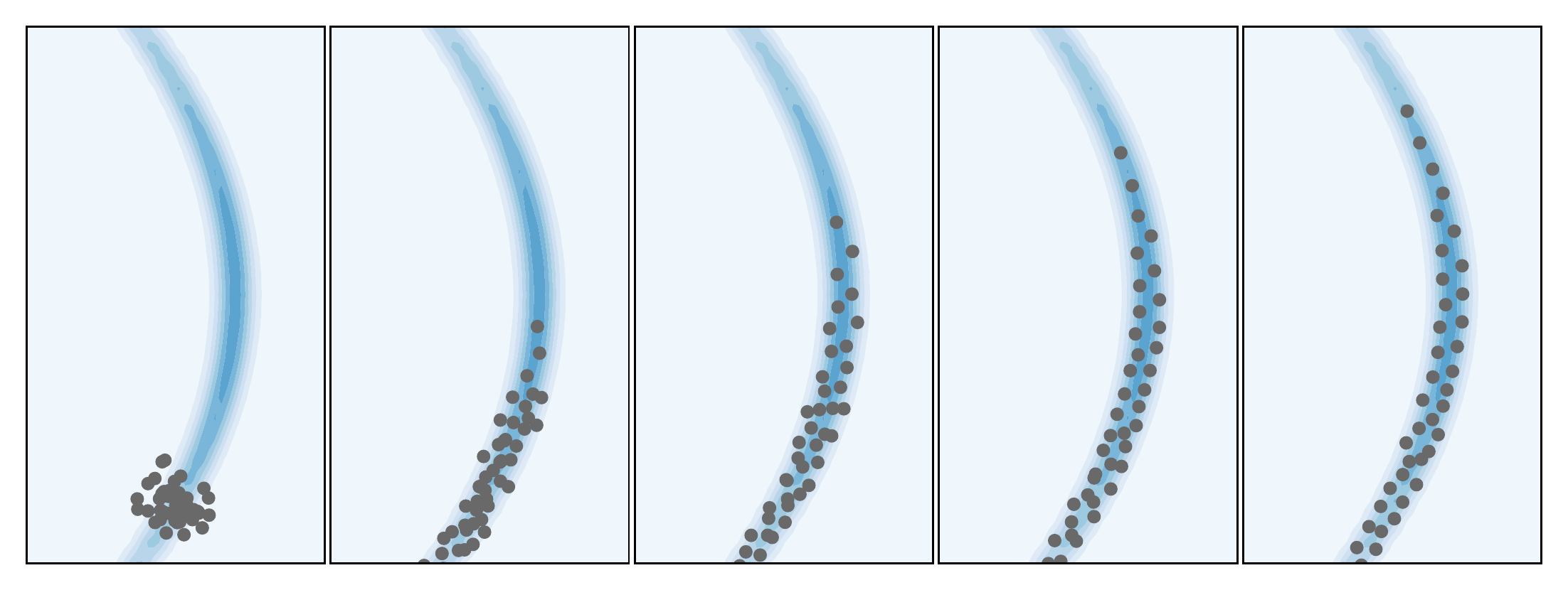} \hspace{-53pt}
  \includegraphics[width=.105\textwidth]{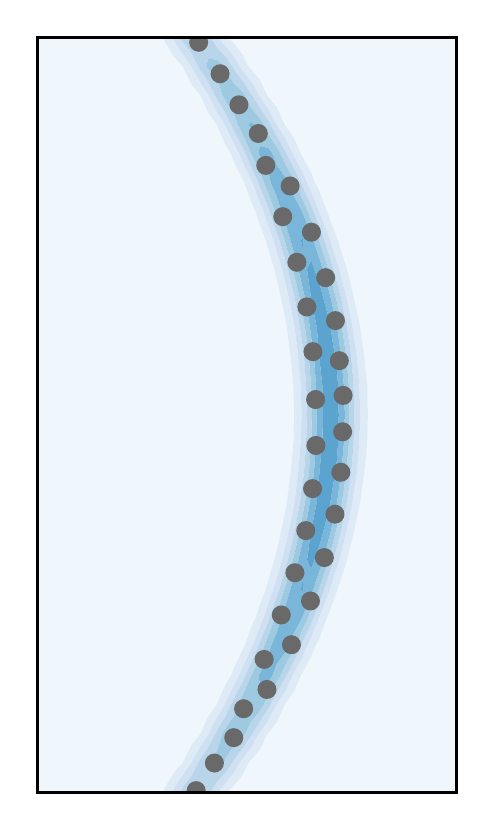} 
  \\[-4.8pt]
  \caption{Dynamics simulation results.
	Rows correspond to Blob, SGHMC, pSGHMC-det, pSGHMC-fGH, respectively.
	All methods adopt the same step size $0.01$, and SGHMC-related methods share the same $\Sigma^{-1} = 1.0$, $C = 0.5$.
	In each row, figures are plotted for every 300 iterations, and the last one for 10,000 iterations. 
	The HE method \cite{liu2019understanding_a} is used for bandwidth selection.
	\vspace{-10pt}
  }
  \vspace{-10pt}
  \label{fig:synth}
\end{figure}

We show in Fig.~\ref{fig:synth} the equivalence of various dynamics simulations, and the advantages of pSGHMC-det and pSGHMC-fGH.
We first find that all methods eventually produce properly distributed particles, demonstrating their equivalence.
For ParVI methods, both proposed methods (Rows 3, 4) converge faster than Blob (Row 1), indicating the benefit of using SGHMC dynamics over LD, where the momentum accumulates in the vertical direction.
For the same SGHMC dynamics, we see that our ParVI versions (Rows 3, 4) converge faster than the vanilla stochastic version (Row 2), due to the deterministic update rule.
Moreover, pSGHMC-fGH (Row 4) enjoys the HE bandwidth selection method \cite{liu2019understanding_a} for ParVIs, which makes the particles neatly and regularly aligned thus more representative for the distribution. 
pSGHMC-det (Row 3) does not benefit much from HE since the density on particles, $q(\theta)$, is not directly used in the dynamics~\origeqref{eqn:psghmcd}.

\subsection{Latent Dirichlet Allocation (LDA)}

\begin{figure}[t]
  \vspace{-4pt}
  \centering
  \subfigure[Learning curve (20 ptcls)]{
	\vspace{-10pt}
	\includegraphics[width=.220\textwidth]{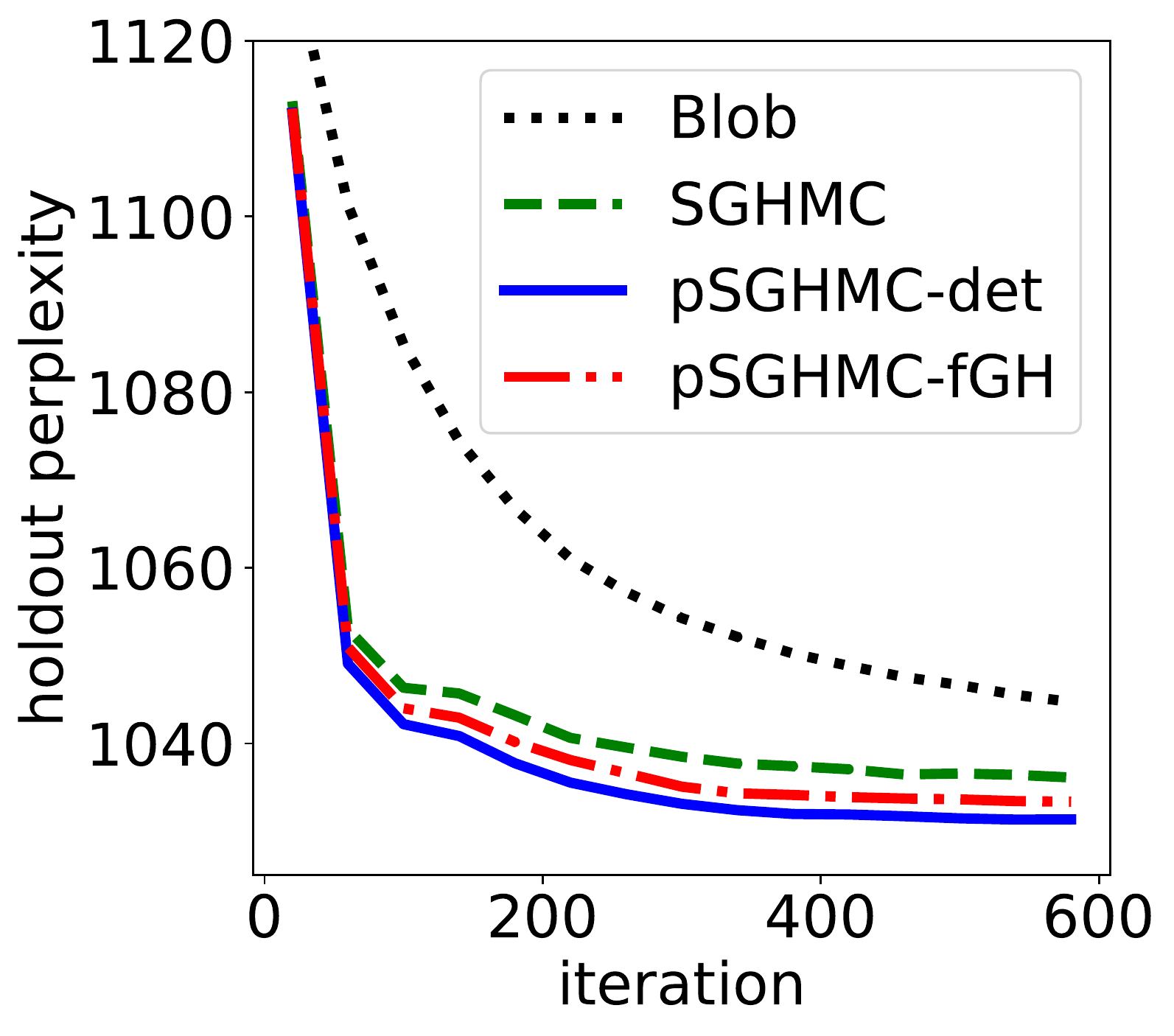}
	\label{fig:lda-lc}
  }
  \hspace{0pt}
  \subfigure[Particle-efficiency (iter 600)]{
	\includegraphics[width=.220\textwidth]{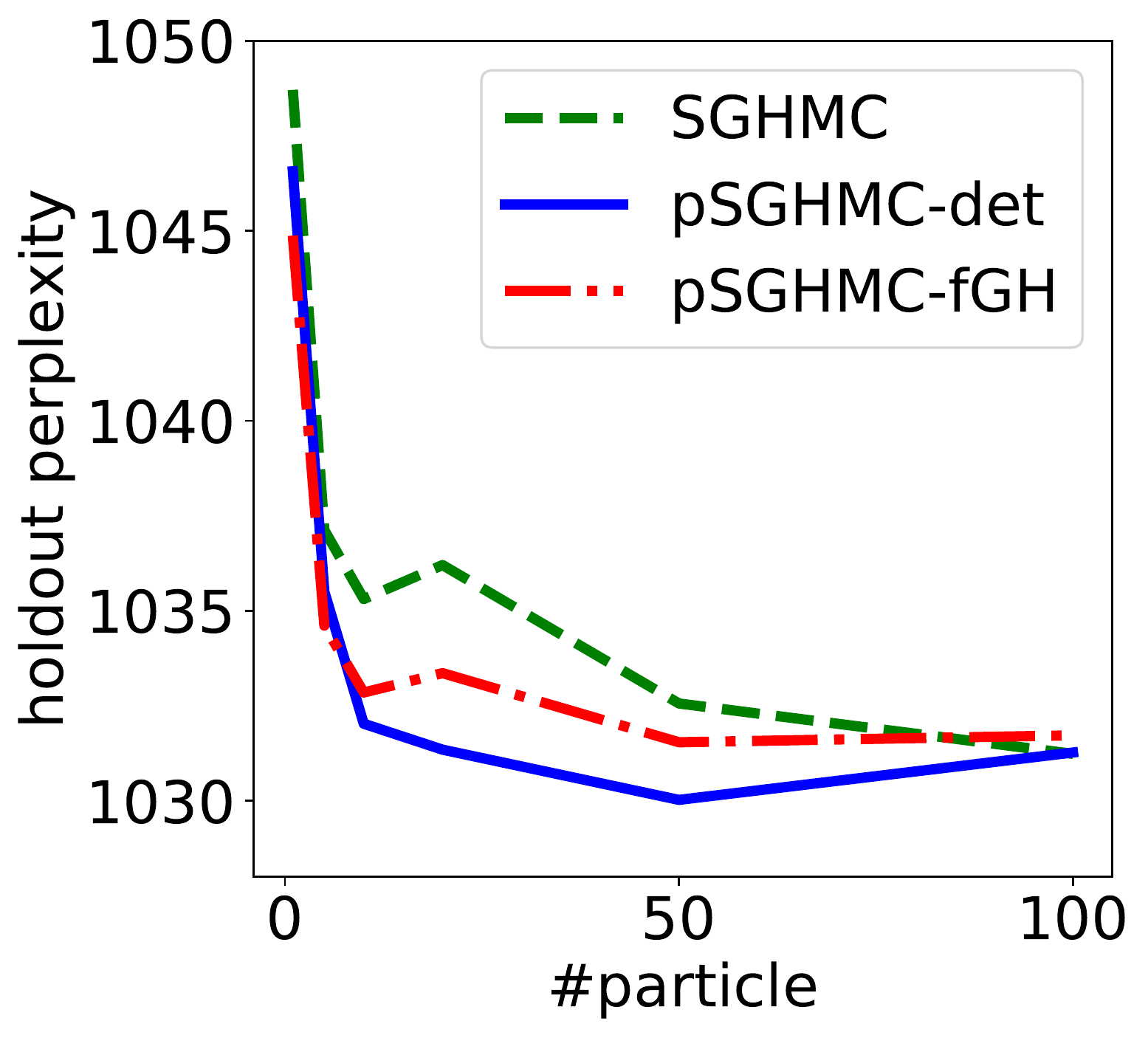}
	\label{fig:lda-ptcl}
  }
  \\[-10pt]
  \caption{Performance on LDA with the ICML data set. 
	Results are averaged over 10 runs.
	All methods share the same step size 0.001 and parameters $\Sigma^{-1} = 300$ and $C = 0.1$.
  }
  \vspace{-10pt}
  \label{fig:lda}
\end{figure}

We study the advantages of our pSGHMC methods in the real-world task of posterior inference for LDA.
We follow the same settings as \citet{liu2019understanding_a} and \citet{chen2014stochastic}.
We see from Fig.~\ref{fig:lda-lc} the saliently faster convergence over Blob, benefited from the usage of SGHMC dynamics in the ParVI field.
Particle-efficiency is compared in Fig.~\ref{fig:lda-ptcl}, where we find the better results of pSGHMC methods over vanilla SGHMC under a same particle size.
This demonstrates the advantage of ParVI simulation of MCMC dynamics, where particle interaction is directly considered to make full use of a set of particles.

\subsection{Bayesian Neural Networks (BNNs)}

\begin{figure}[t]
  \vspace{-4pt}
  \centering
  \subfigure[Learning curve (10 ptcls)]{
	\vspace{-10pt}
	\includegraphics[width=.220\textwidth]{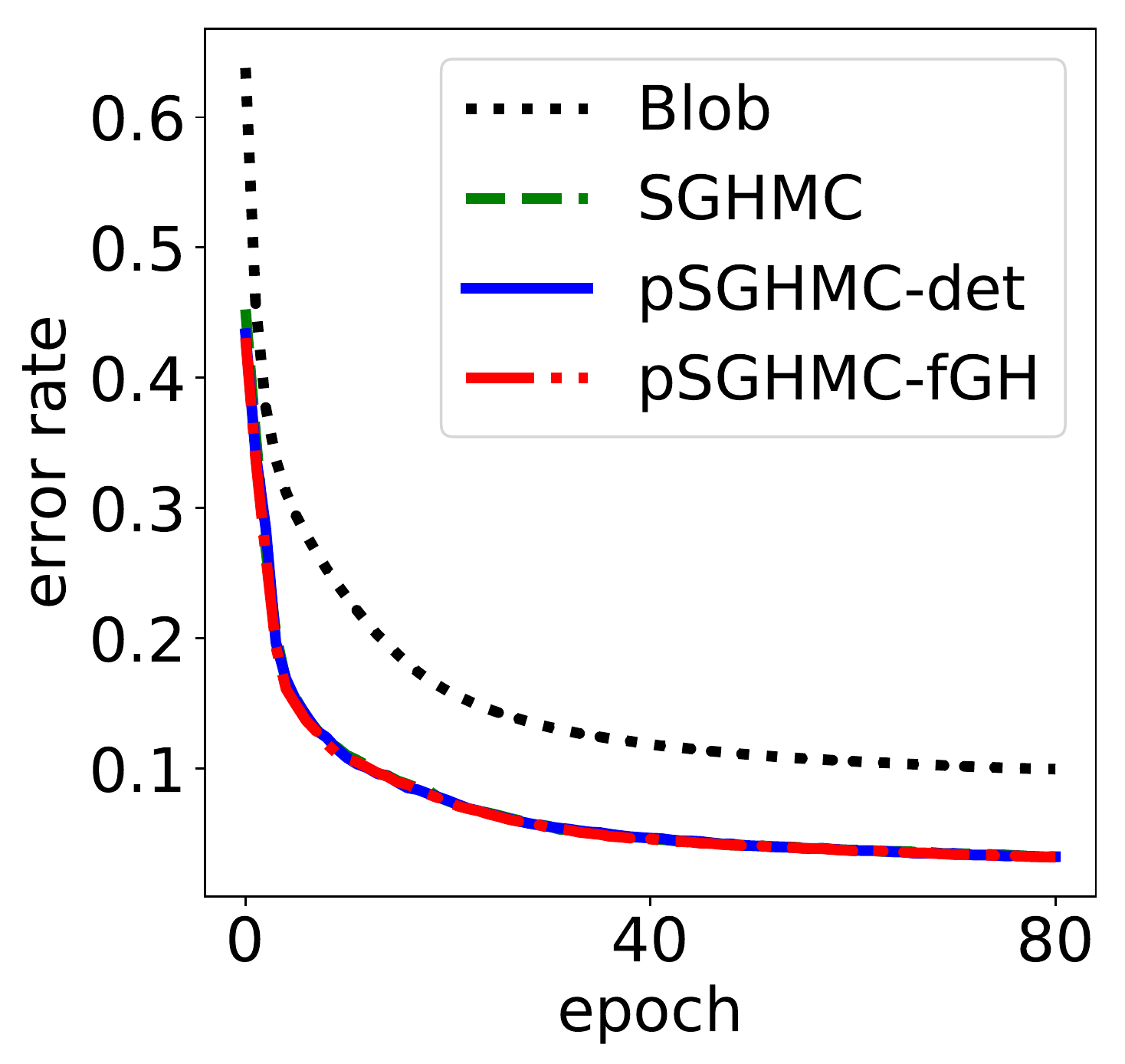}
	\label{fig:bnn-lc}
  }
  \hspace{0pt}
  \subfigure[Particle-efficiency (epch 80)]{
	\includegraphics[width=.231\textwidth]{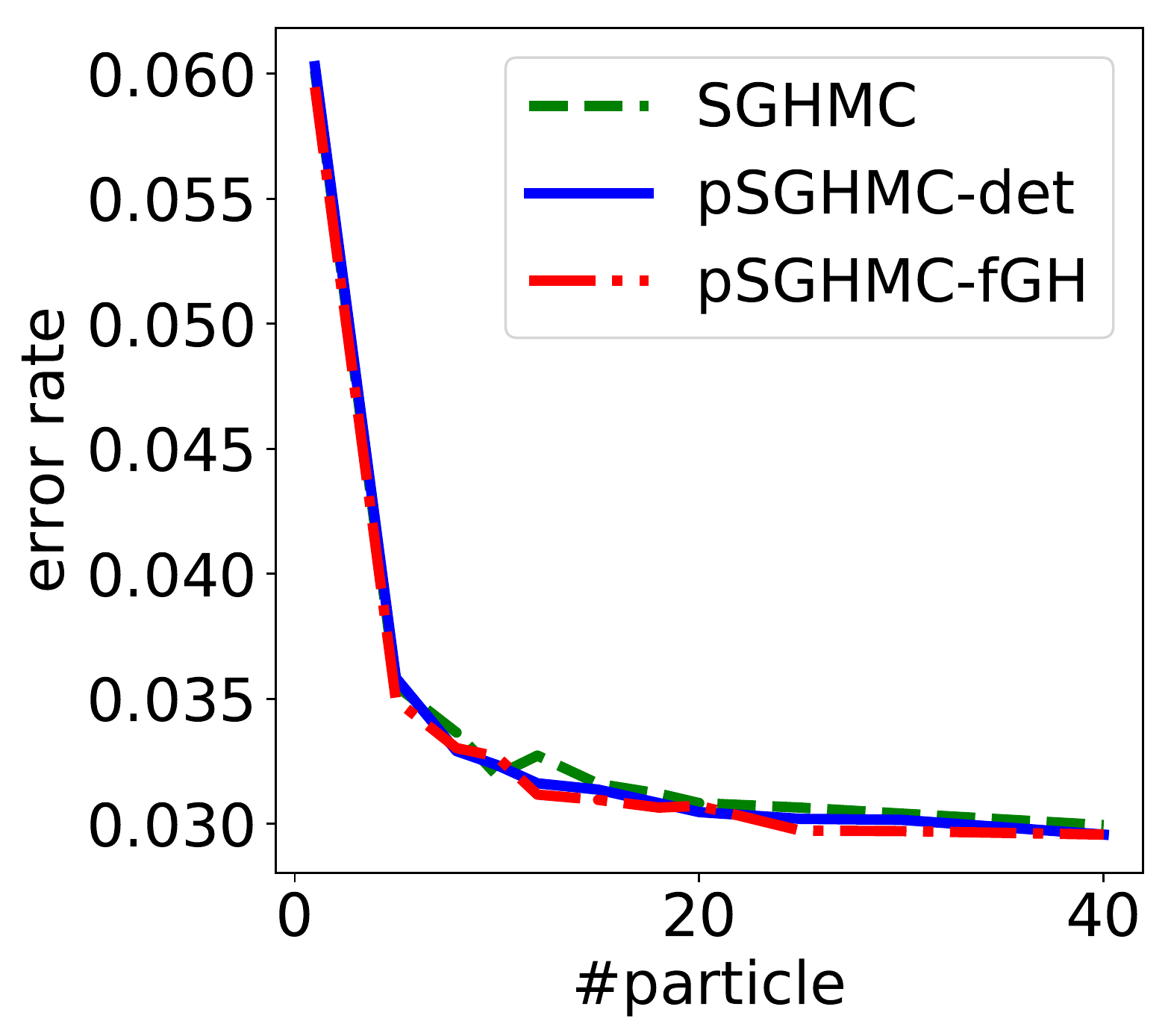}
	\label{fig:bnn-ptcl}
  }
  \\[-10pt]
  \caption{Performance on BNN with MNIST data set.
	Results averaged over 10 runs.
	SGHMC-related methods share parameters.
  }
  \vspace{-16pt}
  \label{fig:bnn}
\end{figure}

We investigate our methods in the supervised task of training BNNs.
We follow the settings of \citet{chen2014stochastic} with slight modification explained in Appendix.
Results in Fig.~\ref{fig:bnn} is consistent with our claim: pSGHMC methods converge faster than Blob due to the usage of SGHMC dynamics.
Their slightly better particle-efficiency can also be observed.

\vspace{-1pt}
\section{Conclusions}
\label{sec:conclusion}
\vspace{-1pt}

We construct a theoretical framework that connects general MCMC dynamics with flows on the Wasserstein space.
By introducing novel concepts, we find that a regular MCMC dynamics corresponds to an fGH flow for an fRP manifold.
The framework gives a clear picture on the behavior of various MCMC dynamics, and also enables ParVI simulation of MCMC dynamics.
We group existing MCMC dynamics into 3 types under the framework and analyse their behavior, and develop two ParVI methods for the SGHMC dynamics.
We empirically demonstrate the faster convergence by more general MCMC dynamics for ParVIs, and particle-efficiency by ParVI simulation for MCMCs.

\section*{Acknowledgments}
This work was supported by the National Key Research and Development Program of China (No. 2017YFA0700904), NSFC Projects (Nos. 61620106010, 61621136008, 61571261), Beijing NSF Project (No. L172037), DITD Program JCKY2017204B064, Tiangong Institute for Intelligent Computing, Beijing Academy of Artificial Intelligence (BAAI), NVIDIA NVAIL Program, and the projects from Siemens and Intel.

\bibliography{refs_thesis}
\bibliographystyle{icml2019}

\newpage

\section*{Appendix}

\subsection*{A. Proofs}
\subsubsection*{A.1. Proof of Lemma~\ref{lem:detmcmc}}

Given the dynamics~\origeqref{eqn:recipe}, the distribution curve $(q_t)_t$ is governed by the Fokker-Planck equation (\eg, \citet{risken1996fokker}):
\begin{align}
  \partial_t q_t = - \partial_i (q_t V^i) + \partial_i \partial_j (q_t D^{ij}),
\end{align}
which reduces to:
\begin{align}
  \partial_t q_t =& - (\partial_i q_t) V^i - q_t (\partial_i V^i) \\
  & {} + q_t (\partial_i \partial_j D^{ij}) + (\partial_i \partial_j q_t) D^{ij} \\
  & {} + (\partial_i q_t) (\partial_j D^{ij}) + (\partial_j q_t) (\partial_i D^{ij}) \\
  =& - (\partial_i q_t) (\partial_j D^{ij} + \partial_j Q^{ij}) - (\partial_i q_t) (D^{ij} + Q^{ij}) \frac{\partial_j p}{p} \\
  & {} - q_t \partial_i \partial_j (D^{ij} + Q^{ij}) - q_t (\partial_i D^{ij} + \partial_i Q^{ij}) \frac{\partial_j p}{p} \\
  & {} - q_t (D^{ij} + Q^{ij}) ( \frac{\partial_i \partial_j p}{p} - \frac{(\partial_i p) (\partial_j p)}{p^2} ) \\
  & {} + q_t (\partial_i \partial_j D^{ij}) + (\partial_i \partial_j q_t) D^{ij} \\
  & {} + (\partial_i q_t) (\partial_j D^{ij}) + (\partial_j q_t) (\partial_i D^{ij}) \\
  =& \hspace{12pt} (\partial_i q_t - \frac{q_t}{p} \partial_i p) (\partial_j D^{ij} - \partial_j Q^{ij}) \\
  & {} - \frac{1}{p} (\partial_i q_t) (\partial_j p) (D^{ij} + Q^{ij}) \\
  & {} - \frac{q_t}{p} (\partial_i \partial_j p) D^{ij} + \frac{q_t}{p^2} (\partial_i p) (\partial_j p) D^{ij} + (\partial_i \partial_j q_t) D^{ij},
\end{align}
where we have used the symmetry of $D$ and skew-symmetry of $Q$ in the last equality:
$(\partial_j p) (\partial_i D^{ij}) = (\partial_i p) (\partial_j D^{ji}) = (\partial_i p) (\partial_j D^{ij})$ and similarly $(\partial_j p) (\partial_i Q^{ij}) = - (\partial_i p) (\partial_j Q^{ij})$;
$\partial_i \partial_j Q^{ij} = \partial_j \partial_i Q^{ji} = -\partial_i \partial_j Q^{ij}$ so $\partial_i \partial_j Q^{ij} = 0$ and similarly $(\partial_i p) (\partial_j p) Q^{ij} = 0$, $(\partial_i \partial_j p) Q^{ij} = 0$.

The deterministic dynamics in the theorem $\ud x = W_t(x) \dd t$ with $W_t(x)$ defined in \eqref{eqn:detmcmc} induces the curve:
\begin{align}
  \partial_t q_t =& - \partial_i (q_t (W_t)^i) \\
  =& - (\partial_i q_t) (W_t)^i - q_t (\partial_i (W_t)^i) \\
  =& - (\partial_i q_t) D^{ij} (\frac{\partial_j p}{p} - \frac{\partial_j q_t}{q_t}) \\
  & {} - (\partial_i q_t) Q^{ij} (\frac{\partial_j p}{p}) - (\partial_i q_t) (\partial_j Q^{ij}) \\
  & {} - q_t (\partial_i D^{ij}) (\frac{\partial_j p}{p} - \frac{\partial_j q_t}{q_t}) \\
  & {} - q_t D^{ij} (\frac{\partial_i \partial_j p}{p} - \frac{(\partial_j p) (\partial_i p)}{p^2} - \frac{\partial_i \partial_j q_t}{q_t} + \frac{(\partial_j q_t) (\partial_i q_t)}{q_t^2}) \\
  & {} - q_t (\partial_i Q^{ij}) \frac{\partial_j p}{p} - q_t Q^{ij} (\frac{\partial_i \partial_j p}{p} - \frac{(\partial_j p) (\partial_i p)}{p^2}) \\
  & {} - q_t (\partial_i \partial_j Q^{ij}) \displaybreak \\
  =& \hspace{12pt} (\partial_i q_t - \frac{q_t}{p} \partial_i p) (\partial_j D^{ij} - \partial_j Q^{ij}) \\
  & {} - \frac{1}{p} (\partial_i q_t) (\partial_j p) (D^{ij} + Q^{ij}) \\
  & {} - \frac{q_t}{p} (\partial_i \partial_j p) D^{ij} + \frac{q_t}{p^2} (\partial_i p) (\partial_j p) D^{ij} + (\partial_i \partial_j q_t) D^{ij},
\end{align}
where we have also applied aforementioned properties in the last equality.
Now we see that the two dynamics induce the same distribution curve thus they are equivalent.

\subsubsection*{A.2. Derivation of \eqref{eqn:barbour}}

Barbour's generator is understood as the directional derivative $(\clA f) (x) = \frac{\ud}{\ud t} F_f (q_t) \Big|_{\substack{q_0 = \delta_x \\ t = 0}}$ on $\clP(\bbR^M)$.
Due to the definition of gradient, this can be written as $(\clA f) (x) = \lrangle{ \grad F_f, \pi_{q_0}( W_0 ) }_{T_{q_0}\clP} = \lrangle{ \grad F_f, W_0 }_{\clL^2_{q_0}}$,
where $\pi_{q_0} (W_0)$ is the tangent vector of the distribution curve $(q_t)_t$ at time $0$ due to Lemma~\ref{lem:detmcmc}, and the last equality holds due to that $\pi_q$ is the orthogonal projection from $\clL^2_q$ to $T_{q} \clP$ and $\grad F_f \in T_{q_0} \clP$ (see Section~\ref{sec:pre-flow-grad}).

Before going on, we first introduce the notion of \emph{weak derivative} (\eg, \citet{nicolaescu2007lectures}, Def.~10.2.1) of a distribution.
For a distribution with a smooth density function $q$ and a smooth function $f \in \clCC(\bbR^M)$, the rule of integration by parts tells us:
\begin{align}
  \int_{\bbR^M} f(x) (\partial_i q(x)) \dd x =& \hspace{10pt} \int_{\bbR^M} \partial_i ( f(x) q(x) ) \dd x \\
  & {} - \int_{\bbR^M} (\partial_i f(x)) q(x) \dd x.
\end{align}
Due to Gauss's theorem (\eg, \citet{abraham2012manifolds}, Thm.~8.2.9), $\int_{\bbR^M} \partial_i ( f(x) q(x) ) \dd x = \lim_{R \to +\infty} \int_{\bbS^{M-1}(R)} ( f(y) q(y) ) v_i(y) \dd y$, where $\bbS^{M-1}(R)$ is the $(M-1)$-dimensional sphere in $\bbR^M$ with radius $R$, $y \in \bbS^{M-1}$, and $v_i$ is the $i$-th component of the unit normal vector $v$ (pointing outwards) on $\bbS^{M-1}(R)$.
Since $f$ is compactly supported and $\lim_{\|x\| \to +\infty} q(x) = 0$, after a sufficiently large $R$, $f(y) q(y) = 0$, so the integral vanishes, and we have:
\begin{align}
  \int_{\bbR^M} f(x) (\partial_i q(x)) \dd x = - \int_{\bbR^M} (\partial_i f(x)) q(x) \dd x, \\
  \forall f \in \clCC(\bbR^M).
\end{align}
We can use this property as the definition of $\partial_i q$ for non-absolutely-continuous distributions, like the Dirac measure $\delta_{x_0}$:
\begin{align}
  \int_{\bbR^M} f(x) (\partial_i \delta_{x_0}(x)) \dd x :=& \int_{\bbR^M} (\partial_i f(x)) \delta_{x_0}(x) \dd x \\
  =& \partial_i f(x_0).
\end{align}

Now we begin the derivation.
Using the form in \eqref{eqn:detmcmc} and noting $q_0 = \delta_{x_0}$, we have:
\begin{align}
  & (\clA f) (x_0) = \lrangle{ \grad F_f, W_0 }_{\clL^2_{q_0}} \\
  =& \bbE_{q_0(x)} [ \lrangle{ \grad f(x), W_0(x) }_{\bbR^M} ] = \bbE_{q_0} [ (\partial_i f) W_0^i ] \\
  =& \bbE_{q_0} \big[ D^{ij} (\partial_i f) \big( \partial_j \log (p/q_0) \big) + Q^{ij} (\partial_i f) (\partial_j \log p) \\
  &\hspace{18pt} + (\partial_j Q^{ij}) (\partial_i f) \big] \\
  =&\hspace{14pt} \big[ D^{ij} (\partial_i f) (\partial_j \log p) \big] (x_0) \\
  & {} - \int_{\bbR^M} \big( D^{ij} (\partial_i f) \big) (x) (\partial_j q_0) (x) \dd x \\
  & {} + \big[ Q^{ij} (\partial_i f) (\partial_j \log p) + (\partial_j Q^{ij}) (\partial_i f) \big] (x_0) \\
  =& \left[ D^{ij} (\partial_i f) (\partial_j \log p) + \frac{1}{p} \partial_j ( p Q^{ij} ) (\partial_i f) \right] (x_0) \\
  & {} + \int_{\bbR^M} \partial_j \big( D^{ij} (\partial_i f) \big) (x) q_0(x) \dd x \\
  =& \left[ D^{ij} (\partial_i f) (\partial_j \log p) + \frac{1}{p} \partial_j ( p Q^{ij} ) (\partial_i f) \right] (x_0) \\
  & {} + \left[ \partial_j \big( D^{ij} (\partial_i f) \big) \right] (x_0) \\
  =& \bigg[ D^{ij} (\partial_i f) (\partial_j \log p) + \frac{1}{p} \partial_j ( p Q^{ij} ) (\partial_i f) \\
  &\hspace{10pt} {} + (\partial_j D^{ij}) (\partial_i f) + D^{ij} (\partial_i \partial_j f) \bigg] (x_0) \\
  =& \left[ \frac{1}{p} \partial_j \big( p (D^{ij} + Q^{ij}) \big) (\partial_i f) + D^{ij} (\partial_i \partial_j f) \right] (x_0) \\
  =& \left[ \frac{1}{p} \partial_j \big( p (D^{ij} + Q^{ij}) \big) (\partial_i f) + (D^{ij} + Q^{ij}) (\partial_i \partial_j f) \right] (x_0) \\
  =& \left[ \frac{1}{p} \partial_j \left[ p \left( D^{ij} + Q^{ij} \right) (\partial_i f) \right] \right] (x_0),
\end{align}
where the second last equality holds due to $Q^{ij} (\partial_i \partial_j f) = 0$ from the skew-symmetry of $Q$.
This completes the derivation.

\subsubsection*{A.3. Proof of Lemma~\ref{lem:klham}}

Noting that the KL divergence $\KL_p(q) = \int_{\clM} \log (q/p) \dd q$ is a non-linear function on $\clP(\clM)$, we need to first find its linearization.
We fix a point $q_0 \in \clP(\clM)$.
\eqref{eqn:gradkl} gives its gradient at $q_0$: $\grad \KL_p (q_0) = \grad \log (q_0 / p)$.
Consider the linear function on $\clP(\clM)$:
\begin{align}
  F: q \mapsto \int_{\clM} \log(q_0/p) \dd q.
\end{align}
According to existing knowledge (\eg, \citet{villani2008optimal}, Ex.~15.10; \citet{ambrosio2008gradient}, Lem.~10.4.1; \citet{santambrogio2017euclidean}, Eq.~4.10), its gradient at $q_0$ is given by:
\begin{align}
  \big( \grad F \big) (q_0) = \grad \left( \left. \frac{\delta F}{\delta q} \right|_{q=q_0} \right),
\end{align}
where $\frac{\delta F}{\delta q}$ is the first functional variation of $F$, which is $\log(q_0/p)$ at $q=q_0$.
Now we find that $\grad F (q_0) = \grad \log(q_0/p) = \grad \KL_p (q_0)$, so $F(q)$ is the linearization of $\KL_p(q)$ at $q=q_0$ and the corresponding $f \in \clCC(\clM)$ in \eqref{eqn:wham} is $\log(q_0/p)$.
Then we have:
\begin{align}
  \clX_{\KL_p} (q_0) = \pi_{q_0}( X_{\log (q_0/p)} ).
\end{align}
Referring to \eqref{eqn:hamfl}, $X_{\log (q_0/p)} = \beta^{ij} \partial_j \log (q_0/p) \partial_i$.
Due to the generality of $q_0$, this completes the proof.

\subsubsection*{A.4. Proof of Theorem~\ref{thm:equiv}}

For a fixed $q \in \clP(\clM)$, two vector fields on $\clM$ produce the same distribution curve if they have the same projection on $T_q \clP(\clM)$, so showing $\pi_q(W) = \clW_{\KL_p}(q)$ is sufficient for showing the equivalence of the two dynamics.
This in turn is equivalent to show $\pi_q(W - \clW_{\KL_p}(q)) = 0_{\clL^2_q}$, or $\divg \big( q (W - \clW_{\KL_p}(q)) \big) = \divg (q 0_{\clL^2_q}) = 0$ (see Section~\ref{sec:pre-flow-grad}). 

We first consider case (b): given an fRP manifold $(\clM, \tgg, \beta)$, we define an MCMC dynamics whose diffusion matrix $D$ and curl matrix $Q$ are the coordinate expressions of the fiber-Riemannian structure $(\tgg^{ij})$ and the Poisson structure $(\beta^{ij})$, respectively.
It is regular, as Assumption~\ref{asm:reg} is satisfied due to properties of $(\tgg^{ij})$ (see \eqref{eqn:fibriem}) and $(\beta^{ij})$ (see Section~\ref{sec:pre-flow-ham}).
Its equivalent deterministic dynamics at $q$ (see Lemma~\ref{lem:detmcmc}) is given by:
\begin{align}
  W^i = \tgg^{ij} \partial_j \log (p/q) + \beta^{ij} \partial_j \log p + \partial_j \beta^{ij}.
\end{align}
So we have:
\begin{align}
  & \divg \big( q (W - \clW_{\KL_p}(q)) \big) \\
  =& \divg \Big( q \big( \tgg^{ij} \partial_j \log (p/q) + \beta^{ij} \partial_j \log p + \partial_j \beta^{ij} \\
  & \hspace{32pt} {} - (\tgg^{ij} + \beta^{ij}) \partial_j \log (p/q) \big) \, \partial_i \Big) \\
  =& \divg \big( q ( \partial_j \beta^{ij} + \beta^{ij} \partial_j \log q ) \partial_i \big) \\
  =& \divg \big( ( q \partial_j \beta^{ij} + \beta^{ij} \partial_j q ) \partial_i \big) \\
  =& \divg \big( \partial_j ( q \beta^{ij} ) \partial_i \big) \\
  =& \partial_i \partial_j ( q \beta^{ij} ) \\
  =& 0,
\end{align}
where the last equality holds due to the skew-symmetry of $(\beta^{ij})$.
This shows that the constructed regular MCMC dynamics is equivalent to the fiber-gradient Hamiltonian flow $\clW_{\KL_p}$ on $\clM$.

For case (a), given any regular MCMC dynamics whose matrices $(D, Q)$ satisfy Assumption~\ref{asm:reg}, we can define an fRP manifold $(\clM, \tgg, \beta)$ whose structures are defined in the coordinate space by the matrices: $\tgg^{ij} := D^{ij}$, $\beta^{ij} := Q^{ij}$.
Assumption~\ref{asm:reg} guarantees that such $\tgg$ is a valid fiber-Riemannian structure and $\beta$ a valid Poisson structure.
On this constructed manifold, we follow the above procedure to construct a regular MCMC dynamics equivalent to the fGH flow $\clW_{\KL_p}$ on it, whose equivalent deterministic dynamics is:
\begin{align}
  W^i = D^{ij} \partial_j \log (p/q) + Q^{ij} \partial_j \log p + \partial_j Q^{ij},
\end{align}
which is exactly the one of the original MCMC dynamics.
This shows that the original regular MCMC dynamics is equivalent to the fGH flow $\clW_{\KL_p}$ on the constructed fRP manifold.

Finally, statement (c) is verified in both cases by the introduced construction.
This completes the proof.

\subsection*{B. Details on Flow Simulation of SGHMC Dynamics}

We first introduce more details on the Blob method, referring to the works of \citet{chen2018unified} and \citet{liu2019understanding_a}.
The key problem in simulating a general flow on the Wasserstein space is to estimate the gradient $u(x) := -\nabla \log q(x)$ where $q(x)$ is the distribution corresponding to the current configuration of the particles.
The gradient has to be estimated using the finite particles $\{ x^{(i)} \}_{i=1}^N$ distributed obeying $q(x)$.
The analysis of \citet{liu2019understanding_a} finds that an estimate method has to make a smoothing treatment, in the form of either smoothing the density or smoothing functions.
The Blob method \cite{chen2018unified} first reformulates $u(x)$ in a variation form:
\begin{align}
  u(x) = \nabla \left( -\frac{\delta}{\delta q} \bbE_{q} [\log q] \right),
\end{align}
then with a kernel function $K$, it replaces the density in the $\log q$ term with a smoothed one:
\begin{align}
  u(x) \approx& \nabla \left( -\frac{\delta}{\delta q} \bbE_{q} [\log (q * K)] \right) \\
  =& -\nabla \log (q * K) - \nabla \left( \frac{q}{(q * K)} * K \right),
\end{align}
where ``*'' denotes convolution.
This form enjoys the benefit of enabling the usage of the empirical distribution: take $q(x) = \hqq(x) := \frac{1}{N} \sum_{i=1}^N \delta_{x^{(i)}}(x)$, with $\delta_{x^{(i)}}(x)$ denoting the Dirac measure at $x^{(i)}$.
The above formulation then becomes:
\begin{align}
  u(x^{(i)}) =& - \nabla_{\! x} \! \log q( x^{(i)} \!) \! \\
  \approx& - \frac{\sum_k \!\! \nabla_{\! x^{(i)}} \! K^{(i,k)}}{\sum_j \! K^{(i,j)}} \!-\! \sum_k \! \frac{\nabla_{\! x^{(i)}} \! K^{(i,k)}}{\sum_j \! K^{(j,k)}},
\end{align}
where $K^{(i,j)} := K(x^{(i)}, x^{(j)})$.
This coincides with \eqref{eqn:blob}.

The vanilla SGHMC dynamics replaces the dynamics $\ud r = - C \nabla_r \log q(r) \dd t$ in \eqref{eqn:psghmcd} with $\ud r = 2 C \dd B_t$ or more intuitively $\ud r = \clN(0, 2 C \dd t)$, where $B_t$ denotes the standard Brownian motion.
The equivalence between these two dynamics can also be directly derived from the Fokker-Planck equation: the first one produces a curve by $\partial_t q_t = - \partial_i \big( q_t (- C^{ij} \partial_j \log q_t) \big) = \partial_i ( C^{ij} \partial_j q_t )$, and the second one by $\partial_t q_t = \partial_i \partial_j ( q_t C^{ij} ) = \partial_i ( C^{ij} \partial_j q_t )$ for a constant $C$, so the two curves coincides.
But dynamics~\origeqref{eqn:psghmcf} cannot be simulated in a stochastic way, since $-\nabla_r \log q(r)$ and $-\nabla_{\theta} \log q(\theta)$ are used to update $\theta$ and $r$, respectively, that is, the correspondence of gradients and variables is switched.
In this case, estimating the gradient cannot be avoided.

Finally, we write the explicit update rule of the proposed methods using Blob with particles $\{ (\theta, r)^{(i)} \}_{i=1}^N$.
Let $K_{\theta}$, $K_r$ be the kernel functions for $\theta$ and $r$, and $\varepsilon$ be a step size.
The update rule for pSGHMC-det in \eqref{eqn:psghmcd} becomes:
\begin{align}
  \begin{cases}
	\theta^{(i)} \asn \theta^{(i)} + \varepsilon \Sigma^{-1} r^{(i)}, \\
	r^{(i)} \asn r^{(i)} + \varepsilon \nabla_{\theta} \log p(\theta^{(i)}) \\
	\hspace{20pt} {} - \varepsilon C \Big( \Sigma^{-1} r^{(i)} + \frac{\sum_k \! \nabla_{\! r^{(i)}} \! K_r^{(i,k)}}{\sum_j \! K_r^{(i,j)}} + \sum_k \! \frac{\nabla_{\! r^{(i)}} \! K_r^{(i,k)}}{\sum_j \! K_r^{(j,k)}} \Big),
  \end{cases}
  \label{eqn:psghmcd-upd}
\end{align}
and for pSGHMC-fGH in \eqref{eqn:psghmcf}:
\begin{align}
  \begin{cases}
	\theta^{(i)} \asn \theta^{(i)} \!+\! \varepsilon \Big( \Sigma^{-1} r^{(i)} \!+\! \frac{\sum_k \! \nabla_{\! r^{(i)}} \! K_r^{(i,k)}}{\sum_j \! K_r^{(i,j)}} \!+\! \sum_k \! \frac{\nabla_{\! r^{(i)}} \! K_r^{(i,k)}}{\sum_j \! K_r^{(j,k)}} \Big), \\
	r^{(i)} \asn r^{(i)} + \varepsilon \nabla_{\theta} \! \log p( \theta^{(i)} \!) \\
	\hspace{20pt} {} - \varepsilon \Big( \frac{\sum_k \! \nabla_{\! \theta^{(i)}} \! K_{\theta}^{(i,k)}}{\sum_j \! K_{\theta}^{(i,j)}} \!+\! \sum_k \! \frac{\nabla_{\! \theta^{(i)}} \! K_{\theta}^{(i,k)}}{\sum_j \! K_{\theta}^{(j,k)}} \Big) \\
	\hspace{20pt} {} - \varepsilon C \Big( \Sigma^{-1} r^{(i)} + \frac{\sum_k \! \nabla_{\! r^{(i)}} \! K_r^{(i,k)}}{\sum_j \! K_r^{(i,j)}} + \sum_k \! \frac{\nabla_{\! r^{(i)}} \! K_r^{(i,k)}}{\sum_j \! K_r^{(j,k)}} \Big),
  \end{cases}
  \label{eqn:psghmcf-upd}
\end{align}
where $K_{\theta}^{(i,j)} := K_{\theta}(\theta^{(i)}, \theta^{(j)})$ and similarly for $K_r^{(i,j)}$.

\subsection*{C. Detailed Settings of Experiments}

\subsubsection*{C.1. Detailed Settings of the Synthetic Experiment}

For the random variable $x = (x_1, x_2)$, the target distribution density $p(x)$ is defined by:
\begin{align}
  \log p(x) =& -0.01 \times \left( \frac{1}{2} (x_1^2 + x_2^2) + \frac{0.8}{2} (25 x_1 + x_2^2)^2 \right) \\
  & {} + \const,
\end{align}
which is inspired by the target distribution used in the work of \citet{girolami2011riemann}.
We use the exact gradient of the log density instead of stochastic gradient.
Fifty particles are used, which are initialized by $\clN \big( (-2,-7), 0.5^2 I \big)$.
The window range is $(-7,3)$ horizontally and $(-9,9)$ vertically.
See the caption of Fig.~\ref{fig:synth} for other settings.

\subsubsection*{C.2. Detailed Settings of the LDA Experiment}

We follow the same settings as \citet{ding2014bayesian}, which is also adopted in \citet{liu2019understanding_a}.
The data set is the ICML data set\footnote{\url{https://cse.buffalo.edu/~changyou/code/SGNHT.zip}} developed by \citet{ding2014bayesian}.
We use 90\% words in each document to train the topic proportion of the document and the left 10\% words for evaluation.
A random 80\%-20\% train-test split of the data set is conducted in each run.

For the LDA model, parameters of the Dirichlet prior of topics is $\alpha = 0.1$.
The mean and standard deviation of the Gaussian prior on the topic proportions is $\beta = 0.1$ and $\sigma = 1.0$.
Number of topics is 30 and batch size is fixed as 100.
The number of Gibbs sampling in each stochastic gradient evaluation is 50.

All the inference methods share the same step size $\varepsilon = 1\e{-3}$.
SGHMC-related methods (SGHMC, pSGHMC-det and pSGHMC-fGH) share the same parameters $\Sigma^{-1} = 300$ and $C = 0.1$.
ParVI methods (Blob, pSGHMC-det and pSGHMC-fGH) use the HE method for kernel bandwidth selection \cite{liu2019understanding_a}.
To match the fashion of ParVI methods, SGHMC is run with parallel chains and the last samples of each chain are collected.

\subsubsection*{C.3. Detailed Settings of the BNN Experiment}

We use a 784-100-10 feedforward neural network with sigmoid activation function.
The batch size is 500.
SGHMC, pSGHMC-det and pSGHMC-fGH share the same parameters $\varepsilon = 5\e{-5}$, $\Sigma^{-1} = 1.0$ and $C = 1.0$, while Blob uses $\varepsilon = 5 \e{-8}$ (larger $\varepsilon$ leads to diverged result).
For the ParVI methods, we find the median method and the HE method for bandwidth selection perform similarly, and we adopt the median method for faster implementation.

\end{document}